\documentclass[runningheads]{llncs}

\usepackage{eccv}

\usepackage{eccvabbrv}

\usepackage{graphicx}
\usepackage{booktabs}

\usepackage[accsupp]{axessibility}  
\usepackage{array}
\usepackage{colortbl}
\usepackage{xcolor}
\usepackage{multirow}
\usepackage{subcaption}
\usepackage{bbding}
\usepackage{pifont}
\makeatletter

\newcommand{\Rmnum}[1]{\expandafter@slowromancap\romannumeral #1@}
\makeatother

\usepackage{hyperref}

\usepackage{orcidlink}

\begin{document}

\title{CPA-Enhancer: Chain-of-Thought Prompted Adaptive Enhancer for Object Detection under Unknown Degradations} 

\titlerunning{Chain-of-Thought Prompted Adaptive Enhancer}

\author{Yuwei Zhang\inst{1}\orcidlink{0000-0002-1370-2912} \and
Yan Wu\inst{1}\orcidlink{0000-0002-8874-8886}\thanks{Corresponding author.} \and
Yanming Liu\inst{2}\orcidlink{0009-0005-1271-1963} \and Xinyue Peng\inst{3}\orcidlink{0009-0009-5512-1831}}

\authorrunning{Y. Zhang~ et al.}

\institute{$^1$Tongji University\quad $^2$Zhejiang University\quad $^3$Southeast University \\
\email{\{yuweizhang,yanwu\}@tongji.edu.cn, oceann24@zju.edu.cn, xinyuepeng@seu.edu.cn}}

\maketitle

\begin{abstract}
Object detection methods under known single degradations have been extensively investigated. However, existing approaches require prior knowledge of the degradation type and train a separate model for each, limiting their practical applications in unpredictable environments. To address this issue, we propose a chain-of-thought (CoT) prompted adaptive enhancer, CPA-Enhancer, for object detection under unknown degradations. Specifically, CPA-Enhancer progressively adapts its enhancement strategy under the step-by-step guidance of CoT prompts, that encode degradation-related information. To the best of our knowledge, it’s the first work that exploits CoT prompting for object detection tasks. Overall, CPA-Enhancer is a plug-and-play enhancement model that can be integrated into any generic detectors to achieve substantial gains on degraded images, without knowing the degradation type priorly. Experimental results demonstrate that CPA-Enhancer not only sets the new state of the art for object detection but also boosts the performance of other downstream vision tasks under unknown degradations. \footnote{The code is available at: \href{https://github.com/zyw-stu/CPA-Enhancer}{https://github.com/zyw-stu/CPA-Enhancer}}
\keywords{Chain-of-thought Prompting \and Adaptive Enhancement  \and Object Detection}
\end{abstract}

\section{Introduction}
Deep learning-based object detection methods, both two-stage\cite{rcnn,fastrcnn,fasterrcnn,maskrcnn,sppnet} and one-stage detectors\cite{ssd, yolov1,yolov2,yolov3,yolov4}, have achieved promising performance in recent years.
However, the detection performance of these methods is severely impaired on degraded images.
To solve this problem, a straightforward strategy involves preprocessing the image through image restoration\cite{msbdn,dcpdn,griddehazenet,dea,ffa,swinir,prenet,restormer,prompt-ir} and enhancement \cite{kind,enlightengan,zero-dce, llflow} methods to improve the image quality. However, a naive preprocessing does not always ensure an improvement in detection performance, leading to suboptimal benefits.
Some researchers resort to unsupervised domain adaptation (UDA) \cite{dayolo, ms-dayolo} and multi-task learning (MTL) \cite{multitask-aet,multitask-dsnet} methods to mitigate this issue. Nonetheless, these methods cannot achieve satisfactory detection results due to the large domain shift.
Other researchers\cite{ia-yolo,gdip-yolo,denet} cascade an enhancement network with a common detector and train them in an end-to-end manner. However, the efficacy of the above methods is constrained to known single degradation scenarios, rendering them inadequate for applying in the real-world unpredictable environments with multiple degradations. 

This leads us to our overarching research question: how can we deploy a unified model that can accurately detect objects under multiple unknown degradations?
CoT prompting is first proposed in natural language processing (NLP), using a sequence of guiding prompts to trigger specific tasks or generate corresponding language model outputs\cite{cot-nlp}.
In computer vision, designing CoT prompts might involve step-by-step analysis of the image content, enabling a model to progressively reason and process visual information\cite{prompt-dehaze,prompt-lowlight,prompt-ir,prompt-video-understanding,prompt-tune1,prompt-tune2,prompt-tune3,prompt-vl-coop,prompt-vl-cocoop}. Based on this cognitive intuition, we believe that CoT prompts could progressively guide the model to adapt its enhancement strategy based on the inferred degradation type.
To this end, we propose a chain-of-thought prompted adaptive enhancer, CPA-Enhancer, to improve the detection performance without knowing the degradation type priorly. Specifically, the key components are the CoT-prompt generation module (CGM) and content-driven prompt block (CPB). 
CGM generates CoT prompts to encode degradation-specific context, while CPB enables the interaction between input features and prompts, allowing the model to adjust its enhancement strategy under the guidance of prompts.

The main contributions of our work are:
1) We propose a chain-of-thought prompted adaptive enhancer, CPA-Enhancer, for object detection under unknown degradations, which is a plug-and-play module that can be trained end-to-end with any common detectors. To the best of our knowledge, it's the first work that exploits CoT prompting for object detection tasks;
2) The key components of CPA-Enhancer are CGM and CPB. CGM generates CoT prompts containing degradation-related information, while CPB helps the model dynamically adjust its enhancement strategies.
3) Experimental results demonstrate that CPA-Enhancer not only achieves favorable detection results against state-of-the-art (SOTA) approaches with well-enhanced features under multiple degradations but also boosts the performance of other downstream vision tasks.

\section{Related Work}
\subsection{Object Detection}
Deep learning-based object detection algorithms can be broadly categorized into single-stage and two-stage approaches\cite{survey-od}.
The mainstream two-stage approaches are the region-based convolutional neural network (R-CNN) series\cite{rcnn,fastrcnn,fasterrcnn,maskrcnn}. The pioneering work of R-CNN~\cite{rcnn} introduces the idea of region proposals, utilizing a selective search for candidate regions and employing convolutional neural network to classify and refine them. Fast R-CNN\cite{fastrcnn} improves speed by sharing convolutional features, and later, Faster R-CNN~\cite{fasterrcnn} introduces region proposal networks that brings faster inference.
Single-stage detection algorithms aim to directly predict bounding box coordinates and class probabilities in a single pass through the network. Examples of single-stage detectors include the you only look once (YOLO) series\cite{yolov1,yolov2,yolov3,yolov4} and the single shot multibox detector (SSD)\cite{ssd}. This paper adopts the YOLOv3\cite{yolov3} model as the baseline detector and enhances its performance under multiple degradations. 

\subsection{Object Detection under Degradation}
Compared to general object detection, detecting objects from degraded images remains a challenging task. 
A straightforward strategy involves preprocessing the image through image restoration\cite{msbdn,dcpdn,griddehazenet,dea,ffa,swinir,prenet,restormer,prompt-ir} and enhancement \cite{kind,enlightengan,zero-dce, llflow} methods to improve the image quality.
Nevertheless, enhancing the image quality does not necessarily guarantee an improvement in detection performance, bringing sub-optimal gains. 
Besides, the complexity of restoration and enhancement models hinders real-time detector performance. 
What's worse, these methods often necessitate supervised learning, requiring a large amount of paired degraded and normal images. 
\cite{ms-dayolo, dayolo} resort to UDA-based methods to address this problem, aiming to acquire silient domain-invariant features from both labeled images in the source domain and unlabeled low-quality images in the target domain.
But, they emphasize aligning data distribution and tend to neglect the latent information present in images. In addition, if the disparity between the two domains is substantial, aligning features from these distinct distributions still remains challenging.
\cite{multitask-aet,multitask-dsnet} apply the MTL-based methods to simultaneously learn visibility enhancement and object detection by sharing the feature extraction layers. 
However, it is hard to tune the parameters to balance the weights between detection and enhancement during training. Besides, the accuracy of MTL-based methods usually decreases on the source domain.%
\cite{ia-yolo,denet} propose an alternative approach that cascades an enhancement network and a common detector, and train them in an end-to-end manner.
However, the aforementioned methods necessitate training separate models for each specific degradation type and require prior knowledge of the input degradation type to apply the appropriate model.

\subsection{Chain-of-thought Prompting}
Prompt learning is first introduced in NLP, using prompts to trigger specific tasks or generate corresponding language model outputs\cite{prompt-gpt3}. 
Due to its notable effectiveness, prompt learning gains prominence in vision-related tasks, such as visual prompt tuning\cite{prompt-tune1,prompt-tune2,prompt-tune3}, low-level vision\cite{prompt-dehaze,prompt-lowlight,prompt-ir}, vision-language models\cite{prompt-vl-coop,prompt-vl-cocoop}, video understanding\cite{prompt-video-understanding} and 3D recognition\cite{prompt-3d-recognition}.
CoT prompting\cite{cot-nlp,fei2023reasoning,lu2023enhancing,eracot} is a recent emerging research direction based on prompt learning, which guides the model through multiple logical steps to enhance the reasoning capabilities of large language models. Additionally, some subsequent vision-related works\cite{CoT-tune,CoT-qa,CoT-seg} are also based on this concept. 
But, it's noteworthy that we are the first to apply CoT prompting to the adaptation of object detection. We substantiate that CoT prompts not only have the potential to generate robust models capable of addressing domain shift challenges but also yield substantial improvement for object detection under image degradations.

\section{Method}
Our goal is to design a pluggable module that enhances important features for down-stream detection tasks without knowing any prior knowledge of the degradation type. 
To this end, we present a chain-of-thought prompted adaptive enhancer, CPA-Enhancer. 
In this section, we first describe the overall architecture for CPA-Enhancer and then discuss its key components in detail.

\subsection{Overview of CPA-Enhancer}
\begin{figure*}[t]
\centering
\begin{minipage}[t]{1\linewidth}
	\includegraphics[width=1\textwidth]{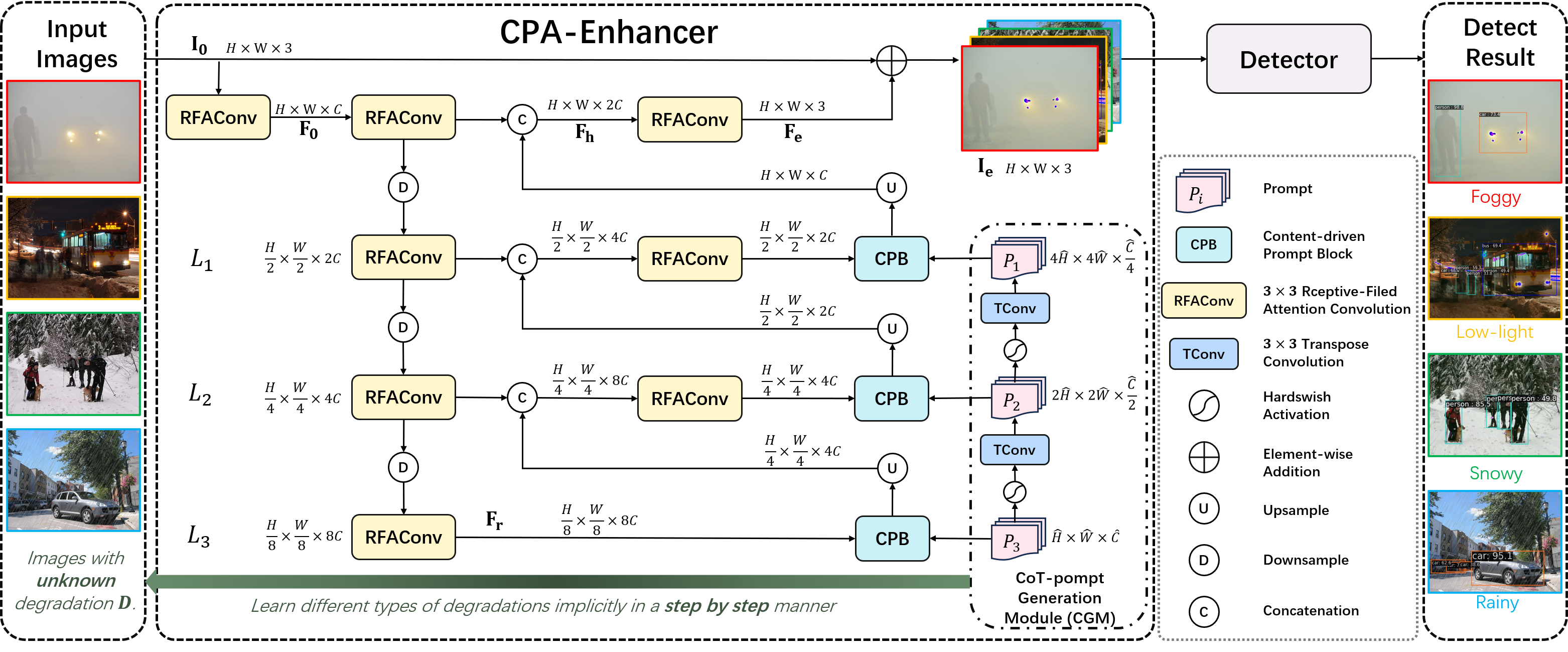}
	\caption{Overview of the proposed CPA-Enhancer.}\label{fig:overall}
\end{minipage}
\end{figure*}
The overall architecture of CPA-Enhancer is illustrated in \cref{fig:overall}. For a given image $\mathbf{I}_{\mathbf{0}} \in \mathbb{R}^{H \times W \times 3}$ with unknown degradation $\mathbf{D}$, CPA-Enhancer initially obtains low-level features $\mathbf{F}_{\mathbf{0}} \in \mathbb{R}^{H \times W \times C}$ by utilizing a receptive-field attention convolution\cite{rfaconv} (RFAConv), where $H \times W$ represents the spatial resolution and $C$ signifies the channels. Subsequently, the embedded features $\mathbf{F}_{\mathbf{0}}$ is fed into a 4-level hierarchical encoder-decoder with each level containing a $3 \times 3$ RFAConv with a stride of $1$. Beginning with features $\mathbf{F}_{\mathbf{0}}$, the encoder progressively decreases the spatial size while increasing the channel capacity through downsampling operations, generating low-resolution latent features $\mathbf{F}_r \in \mathbb{R}^{\frac{H}{8} \times \frac{W}{8} \times 8 C}$. Then, the decoder gradually recovers high-resolution features $\mathbf{F}_{\mathbf{h}} \in \mathbb{R}^{H \times W \times 2C}$ from $\mathbf{F}_{\mathbf{r}}$. Afterward, a RFAConv is applied to generate the enhanced feature $\mathbf{F}_{\mathbf{e}} \in \mathbb{R}^{H \times W \times 3}$, to which the input image is added to obtain the enhanced image: $\mathbf{I}_{\mathbf{e}} = \mathbf{I}_{\mathbf{0}} + \mathbf{F}_{\mathbf{e}}$. Finally, $\mathbf{I}_{\mathbf{e}}$ is used as the input for the YOLOv3 detector.
\begin{figure*}[t]
\centering
\begin{minipage}[t]{1\linewidth}
	\includegraphics[width=1\textwidth]{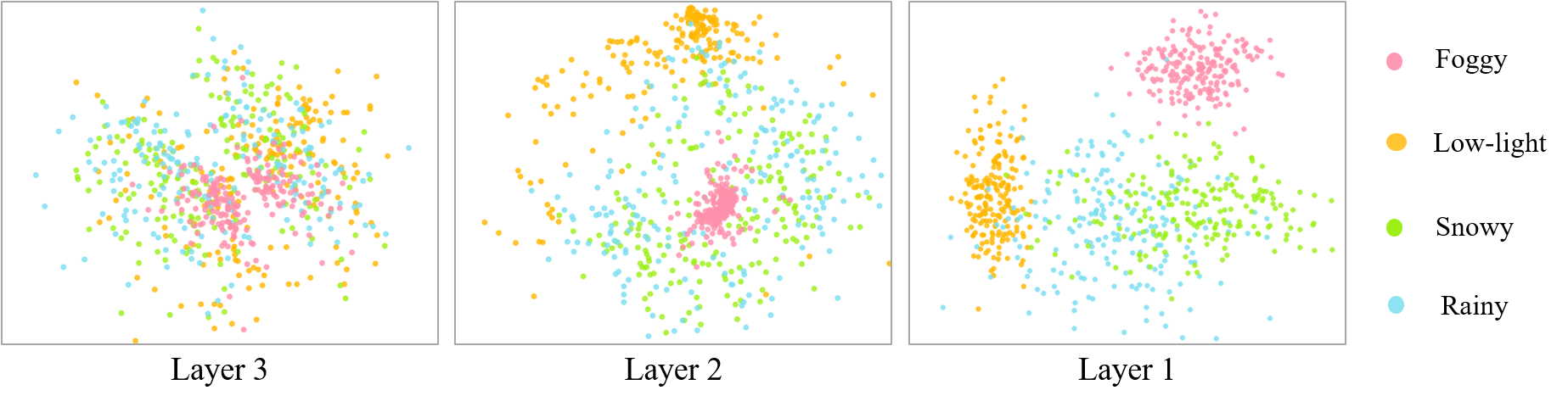}
	\caption{We show t-SNE plots of the degradation embeddings in different decoder layers. It is worth noting that the clusters from Layer 3 to Layer 1 become more and more discriminant, indicating that our network has the ability to distinguish between different types of degradations under the step-by-step guidance of CoT prompts.}\label{fig:tsne}
\end{minipage}
\end{figure*}
To allow the model to dynamically adjust its strategies when faced with different types of degraded images, we propose a chain-of-thought prompt generation Module (CGM) to generates CoT prompts through a series of concatenated layers. To facilitate a guided enhancement process, content-driven prompt block (CPB) is designed to enrich the input features at each decoder level with information about the degradation type. As illustrated in \cref{fig:tsne}, CPA-Enhancer has the capability to progressively process these degradation-related visual cues and helps the model adapt to varying input through a step-by-step guidance of CoT prompts, thus improving its overall robustness and generalization capabilities.
\subsection{CoT-prompt Generation Module (CGM)}
The prompts are learnable parameters used to encode degradation-specific contextual information. The initial prompt in the 3rd Layer is denoted as $\mathbf{P_3} \in \mathbb{R}^{\hat{H}\times  \hat{W} \times \hat {C}}$, where $\hat{H}$, $\hat{W}$, $\hat{C}$ respectively represent the height, width, and channels of the initial prompt. Shown in \cref{fig:overall}, we stack several transpose convolutions\cite{dwconv} layers to generate a sequence of multiscale prompts.
We also apply Hardswish\cite{hardswish} activation functions after all transpose convolutions. The Hardswish activation functions are used to control the flow of information, allowing the network to pass on degradation-related information to the next prompt and ignoring irrelevant ones.
Formally, the procedure of generating CoT Prompts is defined as:
\begin{equation}
    \label{eq:cgm}
    \mathbf{P_i} = \operatorname{Hardswish}(\mathcal{TC}_{3 \times 3}(\mathbf{P_{i+1}})), i\in\{1,2\}
\end{equation}
where $\mathbf{P_i} \in \mathbb{R}^ {2^{(3-i)}\hat{H} \times 2^{(3-i)}\hat{W} \times \frac{\hat{C}}{2^{(3-i)}} } $ denotes the generated prompt in the $i$th layer, $\mathcal{TC}_{k \times k}$ represents $k \times k$ the transpose convolution. The stride of the transpose convolution is set to 2, to double the width and height of the prompts. Hardswish denotes the Hardswish activation function.
In PromptIR\cite{prompt-ir}, each prompt is independent of others and has the same size. By contrast, CGM generates CoT prompts, and the size of each prompt corresponding to the decoding layer is different. This not only helps the model better understand the degradation type in a coherent and step-by-step approach but also assists the model in learning hierarchical representations.
\subsection{Content-driven Prompt Block (CPB)}
\begin{figure*}[t]
\centering
\begin{minipage}[t]{1\linewidth}
	\includegraphics[width=1\textwidth]{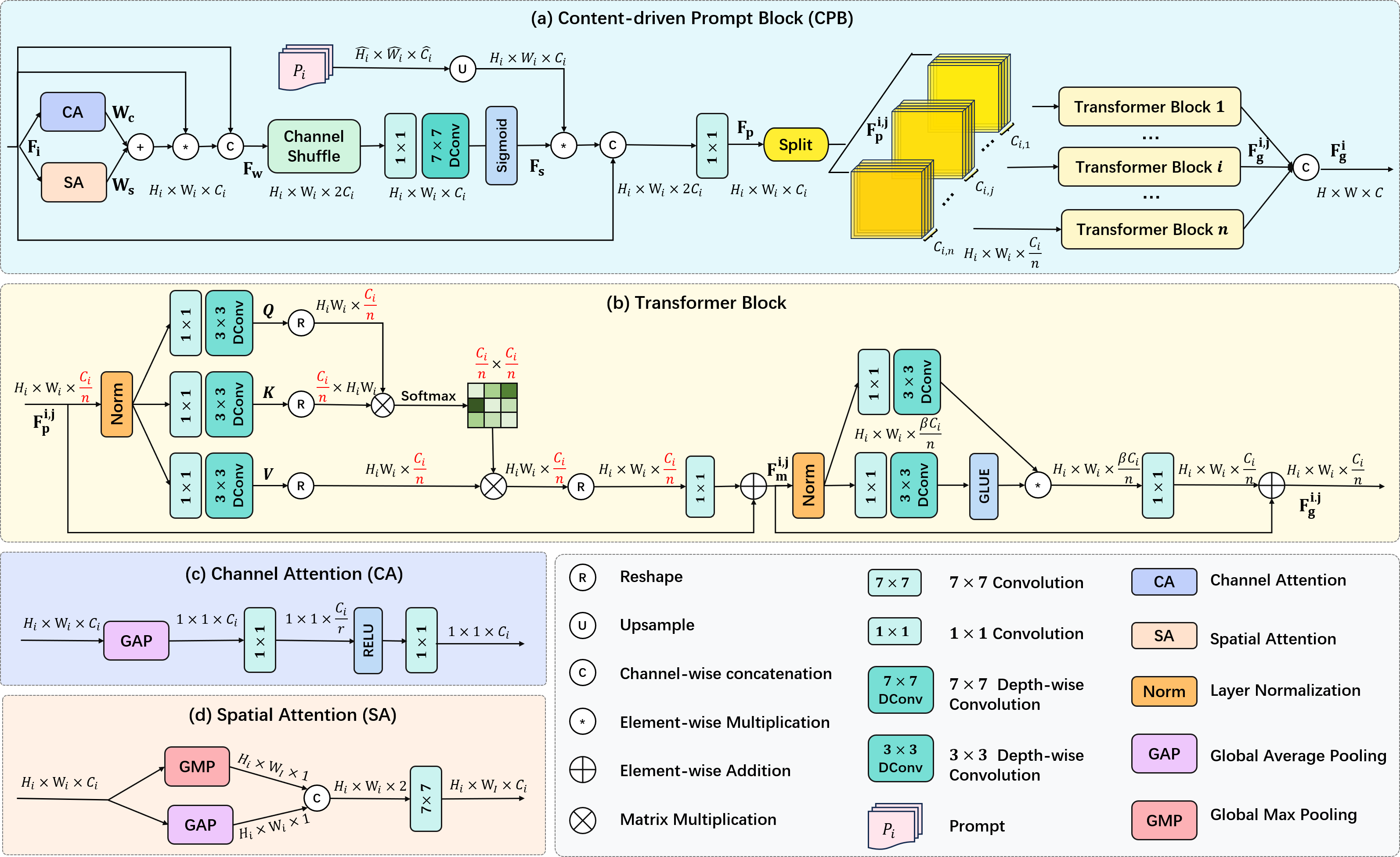}
	\caption{Our proposed content-driven prompt block. }\label{fig:pb}
\end{minipage}
\end{figure*}
CPB is designed to facilitate interaction between the input features $\mathbf{F_i}$ and prompt $\mathbf{P_i}$, enabling the model to adapt its enhancement strategies based on the degradation type. 
To better utilize input content, we calculate a spatial importance map for every single channel\cite{dea}. The attention weights for both channel and spatial dimensions are fully mixed to capture features comprehensively. Next, we split the mixed representations into $n$ equal parts along the channel dimension. Each divided part is fed into an independent transformer block\cite{restormer} that leverages the degradation information encoded in the prompts and transforms the input features. Lastly, all output results are concatenated along the channel dimension. This design can bring the following benefits: 
1) Each part can attend to the relevance of different channels and features, thus increasing the model's expressiveness; 2) It can reduce parameter count and computational complexity. The size of each part of the transposed-attention map is $\mathbb{R}^{\frac{C_i}{n} \times \frac{C_i}{n}}$, since the transformer block calculates the attention in the channel dimension. 3) Each part is calculated independently, enabling parallel processing that can significantly reduce training time.

The detailed procedures of CPB are illustrated in \cref{fig:pb}a. Let $\mathbf{F_i}$ represent the feature of the $i$th decoder layer. Following \cite{ca, sa}, we generate channel-wise attention weight $\mathbf{W_c^i} \in \mathbb{R}^{1 \times 1 \times C_i}$  and spacial-wise attention weight $\mathbf{W_s^i} \in \mathbb{R}^{H_i \times W_i \times C_i}$. $H_i$, $W_i$, $C_i$ respectively represent the height, width, and channels of $i$th decoder layer's output. As depicted  in \cref{fig:pb}c, $\mathbf{W_c^i}$ is derived by \cref{eq:wc}.
\begin{equation}
    \label{eq:wc}
    \mathbf{W_c^i}= \mathcal{C}_{1 \times 1}(\operatorname{RELU}(\mathcal{C}_{1 \times 1}(\operatorname{GAP_c}(\mathbf{F_i}))))
\end{equation}
where $\mathcal{C}_{k \times k}$ denotes $k \times k$ convolution, $\operatorname{RELU}$ represents the RELU activation function, $\operatorname{GAP_c}$ is the global average pooling operation across the spatial dimensions. The first $1\times 1$ convolution reduces the channel dimension from $C$ to $\frac{C}{r}$ (where $r$ is the reduction ratio), and the subsequent $1\times 1$ convolution expands it back to $C$.
As shown in \cref{fig:pb}d, $\mathbf{W_s}$ is generated by \cref{eq:ws}
\begin{equation}
    \label{eq:ws}
     \mathbf{W_s^i}= \mathcal{C}_{7 \times 7}([\operatorname{GAP_s}(\mathbf{F_i}),\operatorname{GMP_s}(\mathbf{F_i})])
\end{equation}
where $\operatorname{GAP_s}$ is the global average pooling operation across channel attention, $\operatorname{GMP_s}$ is the global max pooling operation across channel attention, $[\cdot]$ denotes the channel-wise concatenation operation.
Then $\mathbf{F_s^i}$ is derived by \cref{eq:fs}. 
\begin{equation}
    \label{eq:fs}
    \begin{aligned}
    \mathbf{F_w^i} &=[(\mathbf{W_c^i}\oplus \mathbf{W_s^i}) \odot \mathbf{F_i}, \mathbf{F_i}] \\
    \mathbf{F_s^i}&=\sigma(\mathcal{DC}_{7 \times 7}(CS(\mathbf{F_w^i})))\\
    \mathbf{F_p^i}&= \mathcal{C}_{1\times 1}([\mathbf{F_i}, \operatorname{Rescale}(\mathbf{P_i})\oplus \mathbf{F_s^i}])        
    \end{aligned}      
\end{equation}
where $\odot$ is the element-wise multiplication, $\oplus$ is the element-wise addition, $\sigma$ represents the sigmoid operation, $CS$ denotes the channel shuffle operation\cite{channelshuffle}. $\mathcal{DC}_{k \times k}$ is the depthwise separable convolution with a stride of ${k \times k}$. Different from \cite{dea}, We find through experiments that better results can be achieved by multiplying $\mathbf{F_i}$ with $\mathbf{W_s^i}$ and $\mathbf{W_c^i}$.
$\operatorname{Rescale}$ denotes bilinear interpolation, rescaling the input prompt $\mathbf{P_i}\in \mathbb{R}^{\hat{H_i}\times \hat{W_i} \times \hat {C_i}}$ to the same size as $\mathbf{F_s^i} \in \mathbb{R}^{H_i\times W_i \times C_i}$. 
Then, $\mathbf{F_p^i}$ is splited into $n$ blocks along the channel dimension $C$ by \cref{eq:split}: 
\begin{equation}
    \label{eq:split}
    \mathbf{F_p^{i,j}} = \mathbf{F_p^i}[:,:,(j-1)\frac{C_i}{n}:j\frac{C_i}{n}]
\end{equation}
where $j\in\{1,2,...,n\}$ and $n$ is the total number of transformer blocks. We use an existing Transformer block\cite{restormer} (\cref{fig:pb}b), containing two sequentially connected sub-modules: Multi-Dconv head transposed attention (MDTA), and Gated-Dconv feed forward network (GDFN). 
The $\mathbf{F_p^{i,j}} \in \mathbb{R}^{H_i\times W_i\times \frac{C_i}{n}}$ is fed into MDTA to calculate self-attention weight across channels rather than the spatial dimension, obtaining $\mathbf{F_m^{i,j}}$ by \cref{eq:mdta}.
\begin{equation}
    \label{eq:mdta}
    \mathbf{F_m^{i,j}} = \mathcal{PC} (\mathbf{V} \cdot \sigma(\mathbf{K} \cdot \mathbf{Q} / \alpha))+\mathbf{F_p^{i,j}}
\end{equation}
where $\mathbf{Q}, \mathbf{K}$ and $\mathbf{V}$ respectively represent query, key, and value projections that are obtained by applying $1 \times 1$ point-wise convolutions followed by $3 \times 3$ depth-wise convolutions on the layer normalized input feature maps. $ \mathcal{PC}$ is the point-wise convolution, $\alpha$ denotes a learnable scaling parameter, and $(\cdot)$ represents dot-product interaction. Then, $\mathbf{F_m^{i,j}}$ is passed through GDFN to transform features in a controlled manner, whose procedures can be defined as \cref{eq:gdfn}.
\begin{equation}
\begin{aligned}
    \label{eq:gdfn}
    X_1 &= \operatorname{GELU}(\mathcal{DC}_{3\times 3}(\mathcal{PC}(\operatorname{LN}(\mathbf{F_m^{i,j}}))))\\
    X_2 &= \mathcal{DC}_{3\times 3}(\mathcal{PC}(\operatorname{LN}(\mathbf{F_m^{i,j}}))) \\
    \mathbf{F_g^{i,j}} &= \mathcal{PC}(X_1\odot X_2)+\mathbf{F_m^{i,j}}
\end{aligned}
\end{equation}
where $\operatorname{GELU}$ denotes the GELU activation function, $\operatorname{LN}$ is the layer normalization\cite{layernorm}. Finally, we concatenate $\mathbf{F_g^i}$ along the channel dimension to obtain the final output $\mathbf{F_g^i}$ of CPB by \cref{eq:concat}.
\begin{equation}
    \label{eq:concat}
    \mathbf{F_g^i}=[\mathbf{F_g^{i,1}},...,\mathbf{F_g^{i,j}},...,\mathbf{F_g^{i,n}}], j\in \{1,2,...,n\}
\end{equation}

\section{Experiments}
To demonstrate the effectiveness of CPA-Enhancer, we conduct experiments under two different experimental settings: all-in-one and one-by-one. In the all-in-one setting, we train a unified model that detects objects under four types of degradations, including foggy, low-light, snowy, and rainy. Whereas, for the one-by-one setting, we train separate models for foggy and low-light degradations. 
\subsection{Preparation of Datasets}
There are limited publicly accessible datasets for object detection under degradations.
The RTTS\cite{rtts} dataset is a real-world dataset designed for foggy conditions. It consists of 4,322 natural hazy images annotated with five object classes: person, bicycle, car, bus, and motorcycle. 
The ExDark\cite{exdark} dataset consists of 7,363 real-world images under low-light conditions with ten classes, including bicycle, boat, bottle, bus, car, cat, chair, dog, motorbike, and people. The test set of ExDark is named ExDarkB, and images with the same categories as the RTTS dataset are selected from ExDark to build ExDarkA.  
We use the the aforementioned datasets from real-world scenarios for testing.
Following IA-YOLO\cite{ia-yolo}, we build upon the classic VOC dataset\cite{voc} to generate synthetic datasets. 
We filter out the VOC categories that do not match with RTTS to construct VocNormTrainA (VnA-T) from VOC2007\_trainval and VOC2012\_trainval. VocNormTestA (VnA) is similarly chosen from VOC2007\_test. Analogously, we also construct VocNormTrainB (VnB-T) and VocNormTestB(VnB) by including VOC categories that align with ExDark. \cref{app_1} provides detailed methods for simulating the four types of degradations from clean images. Based on VnA-T, we synthesize our training datasets with four types of degradataions: VocFogTrain (VF-T), VocDarkTrain (VD-T), VocSnowTrain (VS-T), VocRainTrain(VR-T). Similarly, testing datasets are generated from VnA, including VocFogTest (VF), VocDarkTest (VD), VocSnowTest (VS), VocRainTest (VR).
Further, we merge VnA-T, VF-T, VD-T, VS-T,  and VR-T to create VocMultiTrain (VM-T). Considering that VM-T is five times larger than VnA-T, we replicate each image in VnA-T five times, resulting in VocNormTrain (VN-T). Besides, VnA, VF, VD, VS, VR are merged together to construct VM. Following \cite{ia-yolo}, we apply a hybrid strategy that degrades $2/3$ images in VnA-T to build VocFogHybirdTrain (VF-HT).
Similarly, we randomly degrade $2/3$ images in VnB-T to build VocDarkHybirdTrain (VD-HT). The used datasets are summarized in \cref{tab:dataset}. 
\begin{table}[t]
    \centering
    \caption{Overview of the used datasets. The fraction in parentheses represents the proportion of degradation.}
    \begin{tabular}{@{}ccccc@{}}
\toprule
\textbf{Dataset}                   & \textbf{Usage} & \textbf{Type}                                                                        & \textbf{Images} & \textbf{Classes} \\ \midrule
VocNormTestA(VnA)         & test  & no-degradation                                                              & 2734   & 5       \\
VocFogTest(VF)            & test  & synthenic foggy                                                             & 2734   & 5       \\
VocDarkTest(VD)           & test  & synthenic dark                                                              & 2734   & 5       \\
VocSnowTest(VS)          & test  & synthenic snowy                                                             & 2734   & 5       \\
VocRainTest(VR)           & test  & synthenic rainy                                                             & 2734   & 5       \\
RTTS                      & test  & realistic foggy                                                             & 4322   & 5       \\
ExdarkA                   & test  & realistic dark                                                              & 1283   & 5       \\
VocNormTrain(VN-T)        & train & no-degradation                                                              & 40555  & 5       \\
VocMultiTrain(VM-T)       & train & \begin{tabular}[c]{@{}c@{}}synthenic \\ multi-degradation(4/5)\end{tabular} & 40555  & 5       \\
VocMultiTest(VM)          & test  & \begin{tabular}[c]{@{}c@{}}synthenic \\ multi-degradation(4/5)\end{tabular} & 13670  & 5       \\
VocNormTrainA(VnA-T)      & train & no-degradation                                                              & 8111   & 5       \\
VocFogHybridTain(VF-HT)   & train & synthnic foggy(2/3)                                                         & 8111   & 5       \\ \midrule
VocNormTestB(VnB)         & test  & no-degradation                                                              & 3760   & 10      \\
VocDarkTestB(VDB)         & test  & synthenic low-light                                                         & 3760   & 10      \\
ExdarkB                   & test  & realistic low-light                                                         & 2563   & 10      \\
VocNormTrainB(VnB-T)      & train & no-degradation                                                              & 12334  & 10      \\
VocDarkHybridTrain(VD-HT) & train & synthnic low-light(2/3)                                                     & 12334  & 10      \\ \bottomrule
\end{tabular}
    \label{tab:dataset}
\end{table}

\subsection{Implementation Details}
We use the YOLOv3\cite{yolov3} with Darknet-53 backbone as the detector.
All the experiments are performed in Pytorch on RTX 4090 GPU. 
For testing, the image size is fixed at $544\times544$.  In the all-in-one setting, the model is trained on VM-T for 40 epochs. In the one-by-one setting, we train separate models on VF-HT and VD-HT for 240 epochs. Batch size is set to 16, the optimizer uses SGD, and the initial learning rate and the weight decayis set to 0.001 and 0.0005, respectively. We provide complete implementation details in \cref{app_2}.
\subsection{Experimental Results}
The evaluation of performance is based on the mean average precision (mAP) calculated at an intersection over union (IoU) threshold of 0.5, denoted as mAP$_{50}$. The optimal and suboptimal results are displayed in \textcolor{red}{red} and \textcolor{blue}{blue} fonts, respectively. More quantitative and qualitative analysis can be found in \cref{app_3}.

\begin{figure*}[t]
\centering
\begin{minipage}[t]{1\linewidth}
	\includegraphics[width=1\textwidth]{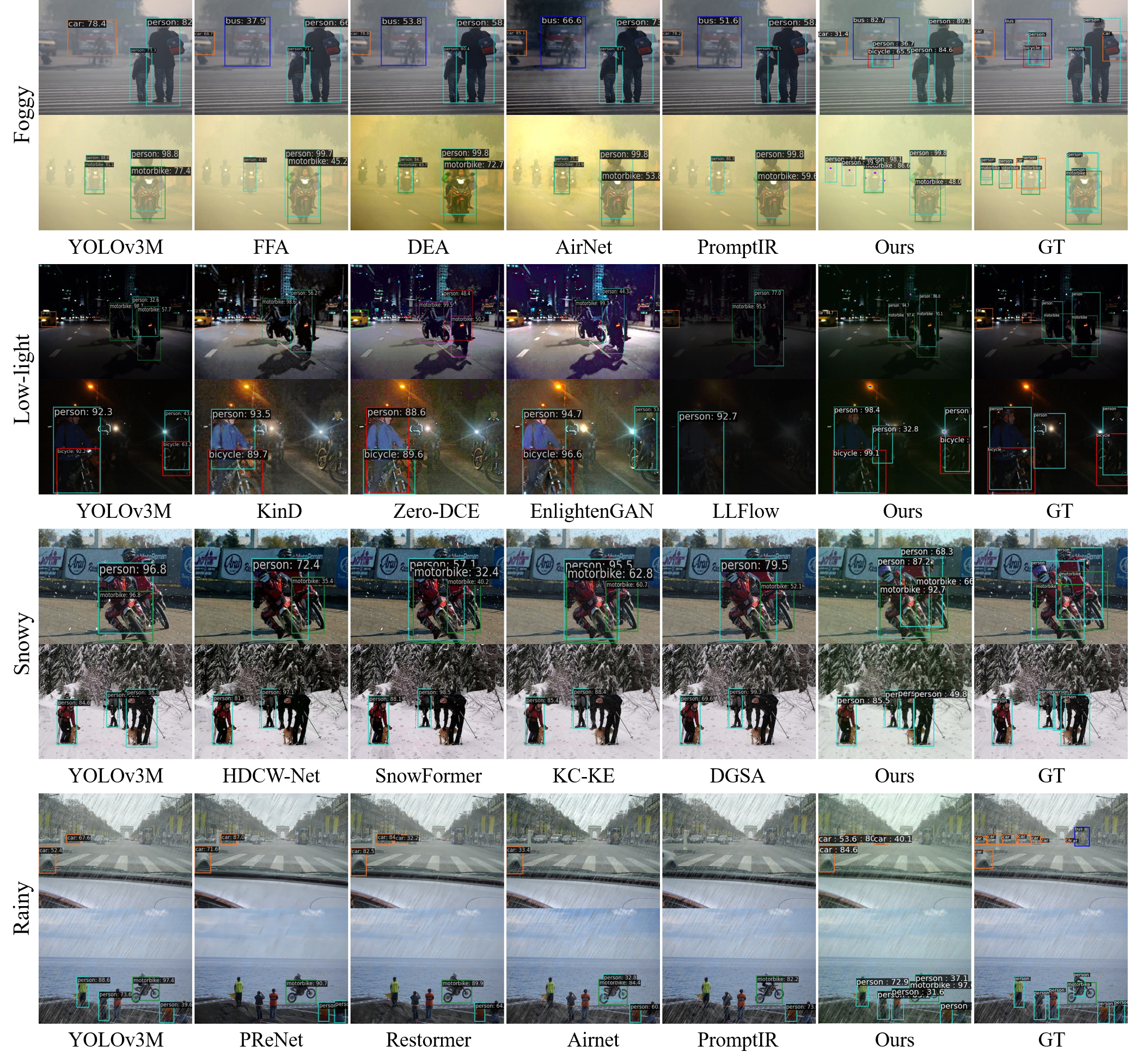}
	\caption{Visual comparisons of CPA-Enhancer under the all-in-one setting.}\label{fig:result}
\end{minipage}
\end{figure*}
\begin{table}[t]
\centering
\caption{Comparisons under the all-in-one setting.}
\begin{tabular}{@{}cc>{\centering\arraybackslash}m{0.8cm}>{\centering\arraybackslash}m{0.8cm}>{\centering\arraybackslash}m{0.8cm}>{\centering\arraybackslash}m{0.8cm}>{\centering\arraybackslash}m{0.8cm}>{\centering\arraybackslash}m{0.8cm}>{\centering\arraybackslash}m{1.85cm}@{}}
\toprule
\multicolumn{2}{c}{\textbf{Methods}} & \textbf{VnA} & \textbf{VF} & \textbf{VD} & \textbf{VS} & \textbf{VR} & \textbf{RTTS} & \textbf{ExDarkA} \\ 
\midrule
\multirow{2}{*}{Baseline} & YOLOv3 (N)\cite{yolov3} & \textcolor{blue}{83.70} & 62.12 & 74.16 & 70.12 & 69.78 & 49.12 & 54.52 \\
 & YOLOv3 (M)\cite{yolov3} & 81.44 & 80.69 & \textcolor{blue}{78.79} & \textcolor{blue}{80.01} & 80.17 & \textcolor{blue}{54.55} & \textcolor{blue}{58.47} \\ 
\midrule
\multirow{2}{*}{Defog} & FFA\cite{ffa} & - & 75.38 & - & - & - & 49.46 & - \\
 & DEA\cite{dea} & - & \textcolor{blue}{81.02} & - & - & - & 50.10 & - \\ 
\midrule
\multirow{4}{*}{\begin{tabular}[c]{@{}c@{}}
Low-light\\ Enhancement\end{tabular}} 
& KinD\cite{kind} & - & - & 74.28& - & - & - & 52.78 \\
& Zero-DCE\cite{zero-dce} & - & - & 76.64 & - & - & - & 53.52 \\
& EnlightenGan\cite{enlightengan} & - & - & 76.42 & - & - & - & 53.16 \\
& LLFlow\cite{llflow} & - & - & 77.08 & - & - & - & 53.90 \\ 
\midrule
\multirow{4}{*}{Desnow} 
& HDCW-Net\cite{snow-hdcw} & - & - & - & 77.82 & - & - & - \\
& SnowFormer\cite{snowformer} & - & - & - & 79.46 & - & - & - \\ 
& KC-KE\cite{snow-twostage} & - & - & - & 79.63 & - & - & - \\ 
& DGSA\cite{snow-dual} & - & - & - & 78.19 & - & - & - \\ 
\midrule
\multirow{2}{*}{Derain} 
& PReNet\cite{penet} & - & - & - & - & 81.32 & - & - \\
& Restormer\cite{restormer} & - & - & - & - & \textcolor{blue}{81.38} & - & - \\ 
\midrule
\multirow{2}{*}{\begin{tabular}[c]{@{}c@{}}
All-in-one\\ Restoration\end{tabular}} 
& Airnet\cite{airnet} & - & 76.20 & - & - & 72.14 & 48.30 & - \\
& PromptIR\cite{prompt-ir} & - & 75.98 & - & - & 73.78 & 50.64 & - \\ 
\midrule
\rowcolor{gray!20} \textbf{Ours} 
& \textbf{CPA-Enhancer} & \textcolor{red}{84.40} & \textcolor{red}{83.32} & \textcolor{red}{80.72} & \textcolor{red}{82.35} & \textcolor{red}{82.30} & \textcolor{red}{56.35} & \textcolor{red}{61.08} \\ 
\bottomrule
\end{tabular}
\label{tab:a_total}
\end{table}

\subsubsection{Multiple Degradation All-in-One Results}
In the all-in-one setting, the baseline methods YOLOv3 (N) and YOLOv3 (M) are derived by training the standard YOLOv3 model\cite{yolov3} on VN-T and VM-T, respectively. In addition, we also compare against a diverse range of methods for pre-processing, including defogging (FFA\cite{ffa}, DEA\cite{dea}, AirNet\cite{airnet}, PromptIR\cite{prompt-ir}), low-light enhancement (KinD\cite{kind}, Zero-DCE\cite{zero-dce}, EnlightenGAN\cite{enlightengan}, LLFlow\cite{llflow}), desnowing (HDCW-Net\cite{snow-hdcw}, SnowFormer\cite{snowformer}, KC-KE\cite{snow-twostage}, DGSA\cite{snow-dual}),  deraining (PReNet\cite{prenet}, Restormer\cite{restormer}, AirNet\cite{airnet}, PromptIR\cite{prompt-ir}).
The pre-trained models of the pre-processing methods are directly applied before YOLOv3 (N).
Reported in \cref{tab:a_total}, we evaluate the performance on seven testing datasets, including VnA, VF, VD, VS, VR, RTTS, and ExDarkA. Although YOLOv3 (M) performs better on images with degradations, it has deterioration on clean images (VnA). Applying image enhancement and restoration methods can only bring limited improvement, and sometimes it can even lead to a decrease in detection performance. This indicates that simply improving the visual quality of images does not necessarily guarantee an improvement in detection performance. Consequently, CPA-Enhancer significantly outperforms all competing methods, demonstrating its effectiveness in object detection under unknown multiple degradations. 

\cref{fig:result} provides visual comparisons. While image enhancement methods can improve image quality to some extent, they can also introduce unwilling noise and artifacts, which can impair detection performance. On the contrary, CPA-Enhancer only enhances the important features for the downstream detection tasks, and adapts its enhancement strategies for different degradation types.
\begin{table}[t]
\centering
\small
\begin{minipage}[t]{0.5\textwidth}
\centering
\small
\caption{Comparisons in the one-by-one\\ setting under the foggy degradation.}
\label{tab:b_fog}
\begin{tabular}{@{}c>{\centering\arraybackslash}m{0.8cm}>{\centering\arraybackslash}m{0.8cm}>{\centering\arraybackslash}m{1cm}@{}}
\toprule
\textbf{Method} & \textbf{VnA} & \textbf{VF} & \textbf{RTTS} \\ 
\midrule
YOLOv3 (NF)\cite{yolov3} & 83.41 & 46.53 & 41.87 \\
YOLOv3 (HF)\cite{yolov3} & 79.06 & 78.87 & 49.45 \\ 
\midrule
GridDehazeNet\cite{griddehazenet} & - & 72.09 & 46.03 \\
DCPDN\cite{dcpdn} & - & 73.38 & 44.43 \\
MSBDN\cite{msbdn} & - & 72.04 & 45.90 \\ 
\midrule
MS-DAYOLO\cite{ms-dayolo} & 81.69 & 65.44 & 42.94 \\
DAYOLO\cite{dayolo} & 80.12 & 66.53 & 44.15 \\ 
\midrule
DSNet\cite{multitask-dsnet} & 71.49 & 81.71 & 49.86 \\ 
\midrule
IA-YOLO\cite{ia-yolo} & 84.05 & 83.22 & 52.36 \\
DE-YOLO\cite{denet} & \textcolor{blue}{84.13} & \textcolor{blue}{83.56} & \textcolor{blue}{53.70} \\
\rowcolor{gray!20}  \textbf{Ours} & \textcolor{red}{86.12} & \textcolor{red}{84.12}& \textcolor{red}{58.55} \\ 
\bottomrule
\end{tabular}
\end{minipage}%
\begin{minipage}[t]{0.5\textwidth}
\centering
\small
\caption{Comparisons in the one-by-one setting under the dark degradation.}
\label{tab:b_dark}
\begin{tabular}{@{}c>{\centering\arraybackslash}m{0.8cm}>{\centering\arraybackslash}m{0.8cm}>{\centering\arraybackslash}m{1.55cm}@{}}
\toprule
\textbf{Method} & \textbf{VnB} & \textbf{VDB} & \textbf{ExDarkB} \\ \midrule
YOLOv3 (ND)\cite{yolov3}                   & 72.22 & 56.34 & 43.02  \\
YOLOv3 (HD)\cite{yolov3}                  & 66.95 & 62.91 & 45.58  \\ \midrule
KinD\cite{kind}                        & -     & 52.57 & 39.22  \\
EnlightenGAN\cite{enlightengan} & -     & 53.67 & 39.42  \\
Zero-DCE\cite{zero-dce}                    & -     & 56.49 & 40.40  \\ \midrule
MS-DAYOLO\cite{ms-dayolo}                   & 72.01 & 58.20 & 44.25  \\
DAYOLO\cite{dayolo}                      & 71.58 & 58.82 & 44.62  \\ \midrule
DSNet\cite{multitask-dsnet}                      & 61.82 & 64.57 & 45.31  \\ \midrule
IA-YOLO\cite{ia-yolo}                     & 72.53 & 67.34 & 49.43  \\
DE-YOLO\cite{denet}                     & \textcolor{blue}{73.17} & \textcolor{blue}{67.81} & \textcolor{blue}{51.51}  \\
\rowcolor{gray!20}  \textbf{Ours}               &   \textcolor{red}{77.02}   & \textcolor{red}{73.09}      &  \textcolor{red}{57.32}       \\ \bottomrule
\end{tabular}
\end{minipage}
\end{table}
\begin{figure*}[h]
\centering
	\includegraphics[width=1\textwidth]{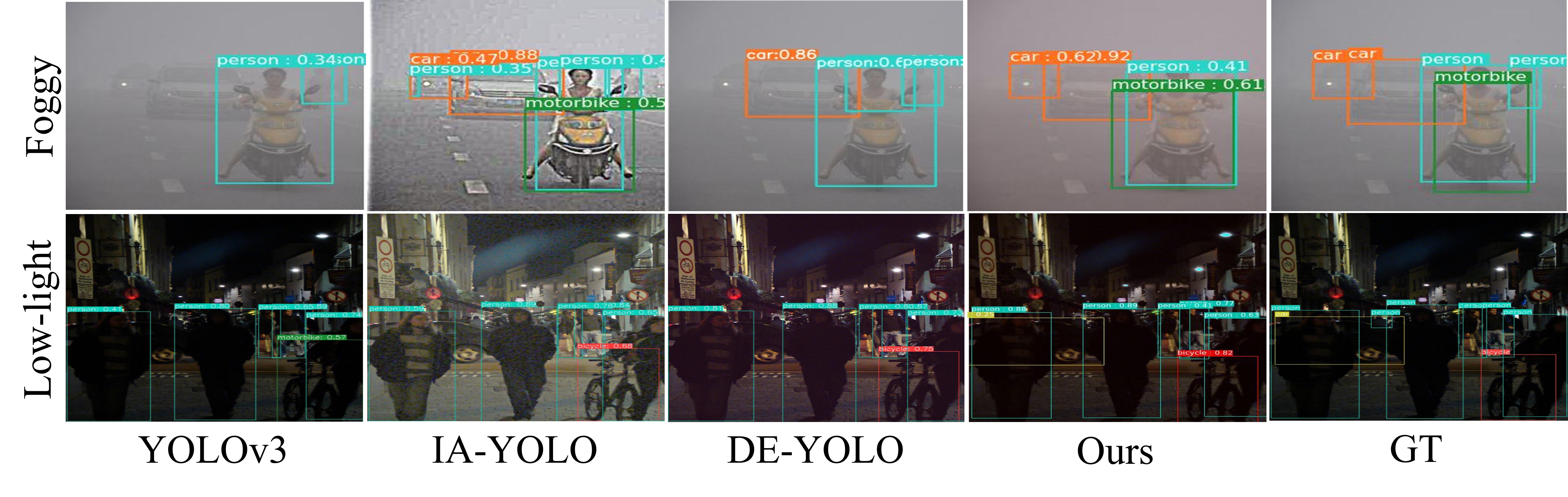}
	\caption{Visual comparison of CPA-Enhancer under the one-by-one setting.}\label{fig:result_dark}
\end{figure*}
\subsubsection{Single Degradation One-by-One Results} 
In the one-by-one setting, we compare CPA-Enhancer with various methods, including image restoration methods\cite{griddehazenet, dcpdn,msbdn} and enhancement methods\cite{kind,enlightengan,zero-dce} as pre-processing, UDA-based methods\cite{ms-dayolo,dayolo}, MTL-based methods\cite{multitask-dsnet}, and methods\cite{ia-yolo,denet} that cascade the enhancement network and YOLOv3. For a fair comparison, we adopt the same experimental settings as DE-YOLO\cite{denet}.
For the foggy condition, the baseline methods YOLOv3 (NF) and YOLOv3 (HF) are created through training the standard YOLOv3 model on VnA-T and VF-HT, respectively. 
Shown in \cref{tab:b_fog}, CPA-Enhancer outperforms all the other methods on all testing datasets by substantial margins. In particular, on the RTTS dataset, CPA-Enhancer surpasses YOLOv3 (HF) and DE-YOLO\cite{denet} by 9.10\% and 4.85\% respectively, achieving mAP$_{50}$ of 58.55\%.  
For the dark condition, the baseline methods YOLOv3 (ND) and YOLOv3 (HD) are derived by training the standard YOLOv3 model on VnB-T and VD-HT, respectively. The quantitative evaluations are reported in \cref{tab:b_dark}. Compared with the current SOTA method DE-YOLO\cite{denet}, the notable increasing performance on VnA ($+3.85\%$ mAP$_{50}$), VDB ($+5.28\%$ mAP$_{50}$), and ExDarkB ($+5.81\%$ mAP$_{50}$) reveals that our method can properly detect objects in low-light environment.
Further, we show the visual comparison between samples generated by the YOLOv3 and the top three methods (IA-YOLO\cite{ia-yolo}, DE-YOLO\cite{denet}, CPA-Enhancer) in \cref{fig:result_dark}. In contrast, our method significantly competes with others in both low-light and foggy conditions.
\subsection{Ablation Studies}
This section ablates key designs and training choices for proposed CPA-Enhancer and further shows its effectiveness on unseen degradation type and other downstream vision tasks. Refer to \cref{app_3} for more details about the experimental results and additional ablation studies.
\begin{table}[t]
\centering
\small
\begin{minipage}[t]{0.6\textwidth}
\centering
\small
\caption{Impact of CGM.}
\label{tab:ablation_prompt}
\begin{tabular}{@{}cccc@{}}
\toprule
\textbf{Method} & \textbf{VM}   & \textbf{RTTS}  & \textbf{ExDarkA} \\ \midrule
w/o prompts       & 81.23          & 54.22          & 58.10           \\
Independent prompts & 81.94          & 55.58         & 59.24          \\
\rowcolor{gray!20} CoT prompts      & \textbf{82.62} & \textbf{56.35} & \textbf{61.08}  \\ \bottomrule
\end{tabular}
\end{minipage}%
\begin{minipage}[t]{0.4\textwidth}
\centering
\small
\caption{Impact of $n$.}
\label{tab:ablation_ns}
\begin{tabular}{@{}cccc@{}}
\toprule
\textbf{n} & \textbf{VM}   & \textbf{RTTS}  & \textbf{ExDarkA} \\ \midrule
2          & 81.80           & 55.82           & 59.88            \\
\rowcolor{gray!20} 4          & \textbf{82.62} & \textbf{56.35} & \textbf{61.08}  \\
8          & 80.86          & 55.06           & 59.32            \\ \bottomrule
\end{tabular}
\end{minipage}%
\end{table}

\begin{table}[t]
\centering
\small
\begin{minipage}[t]{0.48\textwidth}
\centering
\small
\caption{Impact of CPB.}
\label{tab:ablation_cpb}
\begin{tabular}{@{}cccc@{}}
\toprule
\textbf{Method} & \textbf{VM}   & \textbf{RTTS}  & \textbf{ExDarkA} \\ \midrule
SPB          & 81.16          & 55.46          & 59.52           \\
\rowcolor{gray!20} CPB             & \textbf{82.62} & \textbf{56.35} & \textbf{61.08}  \\ \bottomrule
\end{tabular}
\end{minipage}
\begin{minipage}[t]{0.48\textwidth}
\centering
\small
\caption{Performance on noisy datasets.}
\label{tab:ablation_noise}
\begin{tabular}{@{}cccc@{}}
\toprule
\textbf{Method}             & $\sigma = 15$ & $\sigma = 25$ & $\sigma = 50$ \\ \midrule
YOLOv3\cite{yolov3}    & 80.27  & 79.81     & 78.94              \\
\rowcolor{gray!20} \textbf{Ours} & \textbf{82.34}  & \textbf{81.95} & \textbf{80.62} \\ \bottomrule
\end{tabular}
\end{minipage}
\end{table}
\begin{table}[hb]
\centering
\caption{Comparisons on the ACDC validation set.}
\begin{tabular}{@{}c>{\centering\arraybackslash}m{2.8cm}>{\centering\arraybackslash}m{2.8cm}>{\centering\arraybackslash}m{2.8cm}@{}}
\toprule
\textbf{Method} & Refign\cite{refign} (DAFormer\cite{seg-daformer})  & CPA-Enhancer (DeeplabV3+\cite{seg-deeplabv3}) & CPA-Enhancer (SegFormer\cite{segformer}) \\ \midrule
\textbf{mIoU}   & 65.0                        & 69.3                      & \textbf{75.1}                     \\ \bottomrule
\end{tabular}
\label{tab:seg}
\end{table}
\subsubsection{Impact of CGM.}
To examine the effect of CGM, we conduct this ablation experiment on VM, RTTS\cite{rtts} and ExdarkA\cite{exdark}.
\cref{tab:ablation_prompt} shows that models with prompts outperform the method without prompts, indicating that prompts help the model better adapt to the different degradation types.
Further, the CoT-prompt design yields performance gains over the independent prompt design. 
It demonstrates that the model's ability can be enhanced by 
incorporating context and dependencies between prompts and step-by-step analysis.
\subsubsection{Impact of CPB.}
To assess the impact of CPB, we design a Simple Prompt Block (SPB) for comparison. SPB only performs element-wise multiplication between $\mathbf{F_i}$ and the prompt $\mathbf{P_i}$. The resulting output is then concatenated with $\mathbf{F_i}$ along the channel dimension and passed through a $1\times 1$ convolution to get the final output. As reported in \cref{tab:ablation_cpb}, the notable improvement on VM (+1.46\% mAP$_{50}$), RTTS (+0.89 \%mAP$_{50}$), and ExDarkA (+1.56\% mAP$_{50}$) reveals that CPB is more powerful than SPB in facilitating interaction between $\mathbf{F_i}$ and $\mathbf{P_i}$.
\begin{figure}[h]
  \centering
  \begin{subfigure}[b]{0.45\textwidth}
    \includegraphics[width=\textwidth]{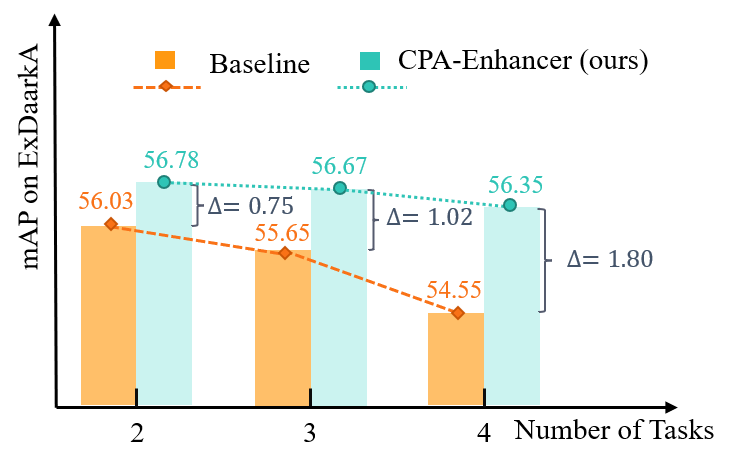}
    \caption{Average preformance on ExDarkA.}
  \end{subfigure}
  \begin{subfigure}[b]{0.45\textwidth}
    \includegraphics[width=\textwidth]{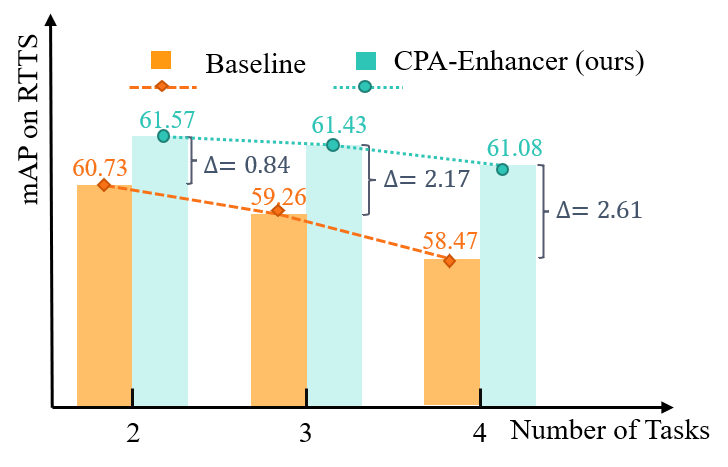}
    \caption{Average preformance on RTTS.}
  \end{subfigure}
  \caption{Impact of the number of tasks.}
  \label{fig:task_num}
\end{figure}

\subsubsection{Impact of $n$.}
The key parameter of our model is the number $n$ of split blocks in the CPB module. As Listed in \cref{tab:ablation_ns}, the best detection performance is achieved when $n$ is set to 4.

\subsubsection{Impact of task numbers.} We train CPA-Enhancer and the baseline YOLOv3\cite{yolov3} by different numbers of tasks and evaluate the average performance on RTTS and ExDarkA.
\cref{fig:seg} shows that with an increasing number of tasks, CPA-Enhancer experiences a smaller mAP$_{50}$ decline compared to YOLOv3, indicating its superior performance stability and robustness when handling multiple tasks.
\subsubsection{Generalization to unseen degradation types.}
To synthetic noisy datasets, we add the Gaussian noise multiplied by a certain coefficient $\sigma\in\{15,25,50 \}$ to the clean images from VnA. Displayed in \cref{tab:ablation_noise}, CPA-Enhancer trained in the all-in-one setting significantly outperforms the baseline YOLOv3\cite{yolov3} under different noise levels, showcasing its excellent generalization capabilities.

\subsubsection{Performance on other downstream visions tasks.}
\begin{figure*}[t]
\centering
	\includegraphics[width=1\textwidth]{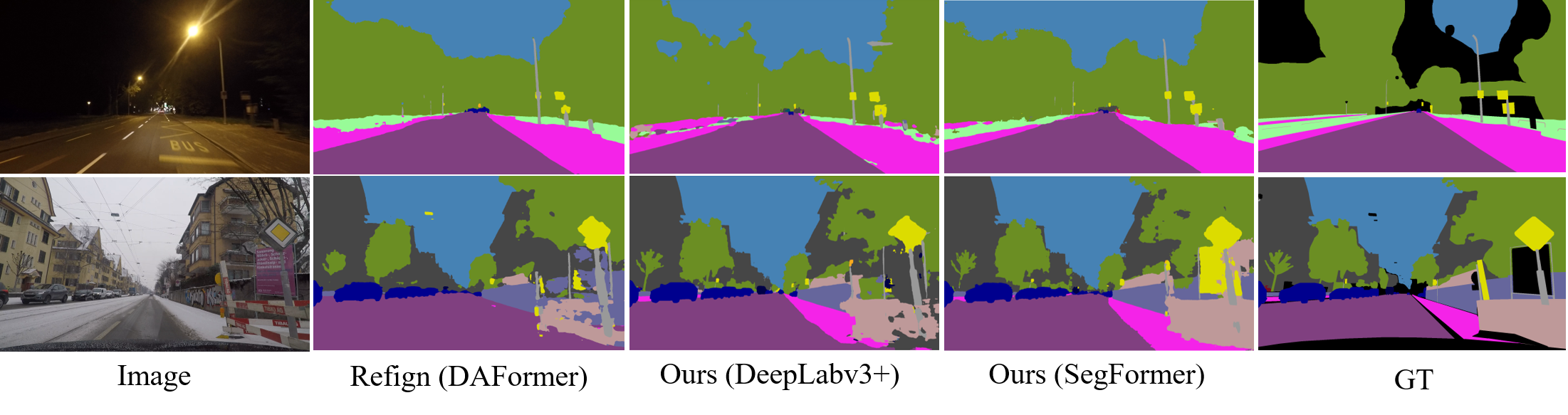}
	\caption{Qualitative segmentation results on the ACDC validation set.}\label{fig:seg}
\end{figure*}

We enhance semantic segmentation models under challenging conditions (fog, night, rain, snow) by cascading them with CPA-Enhancer. Training end-to-end on the ACDC\cite{acdc} training set, our approach outperforms Refign\cite{refign} (based on DAFormer\cite{seg-daformer}) on the ACDC validation set. Evaluation using mean intersection over union (mIoU) shows significant improvements, with CPA-Enhancer boosting DeepLabV3+\cite{seg-deeplabv3} and SegFormer\cite{segformer} to mIoU values of 69.3\% and 75.1\% (see \cref{tab:seg}). \cref{fig:seg} illustrates our model's capability to produce more precise segmentation results.

\subsection{Efficiency Analysis}
CPA-Enhancer introduces 3M trainable parameters and requires only an additional 3 ms compared to the YOLOv3 baseline when detecting a $544\times 544\times 3$ resolution image on a single RTX 4090 GPU. 
Although our CPA-Enhancer takes about 1 ms more per image than DE-YOLO\cite{denet}, it shows remarkable improvement on all the testing datasets. \cref{app_4} presents detailed efficiency analysis.

\section{Conclusion}
Current object detection methods for degraded images necessitate prior knowledge of the degradation type and the training of specific models for each type, which is unsuitable for real-world applications in unpredictable environments.
To this end, we propose the CPA-Enhancer for object detection under unknown multiple degradations. The generated CoT prompts enable the model to adapt its enhancement strategies based on the type of degradations. To the best of our knowledge, it's the first work that exploits CoT prompting for object detection tasks. Experimental results shows that CPA-Enhancer boosts the performance of downstream vision tasks, exceeding SOTA methods.  
In the future, we aim to create a more universal model that covers a wider range of degradations.

%
%
\bibliographystyle{splncs04}
\bibliography{main}
\clearpage
\appendix
\textbf{Appendix Overview:} 
Firstly, \cref{app_1} introduces the methods for generating five types of synthetic datasets, including foggy (see \cref{data_fog}), low-light (see \cref{data_dark}), snowy (see \cref{data_snow}), rainy (see \cref{data_rain}), and noisy (see \cref{data_noise}) datasets. Secondly, complete implementation details of our experiments are provided in \cref{app_2}. Thirdly, \cref{app_3} presents additional quantitative and qualitative experimental results and analyses, as well as supplementary ablation experiments. \cref{app_31} and \cref{app_32} further investigate the impact of the number of tasks and the initial prompt size. \cref{app_spb} illustrates the simple prompt block (SPB). \cref{app_33} presents direct comparisons with the previous state-of-the-art method in the low-light condition. \cref{app_34} provides the detailed performance of our CPA-Enhancer on the semantic segmentation task. \cref{app_35} shows more visual comparisons for the all-in-one setting.
Finally, \cref{app_4} offers detailed efficiency analysis.
\section{Dataset Preparation}
\label{app_1}
\subsection{Synthetic Foggy Dataset}
\label{data_fog}
To synthesize foggy datasets VocFogTrain (VF-T) and VocFogTest (VF), the images from VnA-T and VnA are degraded by employing the atmospheric scattering model (ASM)\cite{asm} respectively: 
\begin{equation}
    \label{eq:foggy}
    \mathbf{I_f}=\mathbf{I} e^{-\beta \mathbf{d}} +A(1-e^{-\beta \mathbf{d}} )
\end{equation}
where $\mathbf{I_f}$ represents the synthetic foggy image, $A$ denotes the global atmospheric light and is fixed at 0.5 in our experimental setup, $\beta = 0.05+0.01*i$, $i$ is a random integer number that ranges from 0 to 9. $d$ is defined as $d=-0.04 * \rho+\sqrt{\max (h,w)}$, where $\rho$ is the Euclidean distance from the current pixel to the central pixel, $h$ and $w$ are the height and width of the image, respectively.

\subsection{Synthetic Low-light Dataset}
\label{data_dark}
To synthesize dark datasets VocDarkTrain (VD-T) and VocDarkTest (VD), the original images in VnA-T and VnA are degraded by gamma transformation:
\begin{equation}
    \label{eq:dark}
    \mathbf{I_d} = \mathbf{I}^{\gamma}
\end{equation}
where the value of $\gamma$ is randomly sampled from a uniform distribution within the range $[1.5, 5]$, and $x$ denotes the input pixel intensity. Similarly, VocDarkTestB (VDB) is obtained by \cref{eq:dark} from VnB.

\subsection{Synthetic Snowy Dataset}
\label{data_snow}
To synthesize snowy datasets VocSnowTrain (VS-T) and VocSnowTest (VS), we overlap the medium snow mask\cite{desnownet} to the images from VnA-T and VnA:
\begin{equation}
   \label{eq:noisy}
   \mathbf{I_s} = \mathbf{I}+M_s
\end{equation}
where $M_s$ denotes the generated snow mask via PhotoShop.

\subsection{Synthetic Rainy Dataset}
\label{data_rain}
To synthesize rainy datasets VocRainTrain (VR-T) and VocRainTest (VR), we take randomly generated Gaussian noise and apply rotation, Gaussian blurring, and filtering to create a raindrop effect. Then, we overlay the raindrop effect on the original image from VnA-T and VnA by creating a 4-channel image with an alpha channel that controls the transparency of the raindrops:
\begin{equation}
    \label{eq:rainy}
    \mathbf{I_r}[:,:,i]=\mathbf{I}[:,:,i]\times(1-R)+\beta\times\mathbf{I}[:,:,i]
\end{equation}
where $i\in \{0,1,2\}$ represents the $i$th channel of the image. $R$ denotes the generated rain effect. $\beta$ is a weighting parameter that is set to 0.8 in our experiments. It controls the blending degree of the raindrop effect with the original image.
\
\subsection{Synthetic Noisy Dataset}
\label{data_noise}
To synthesize noisy datasets VocNoiseTrain (VN-T) and VocNoiseTest (VN), we add the generated Gaussian noise to the images from VnA-T and VnA:
\begin{equation}
   \label{eq:noisy}
    \mathbf{I_n} = \mathbf{I}+N\times {\sigma}
\end{equation}
where $N$ represents randomly generated noise with the same size as the input image $I$, $\sigma$ denotes a specified standard deviation to control the noise level. 
\section{Implementation Details}
\label{app_2}
We use the YOLOv3\cite{yolov3} with Darknet-53 backbone as the detector and mmdetection\cite{mmdetection} is utilized to achieve our model. 
All the experiments are performed on a single RTX 4090 GPU. In the default setting, the initial prompt size is $32\times 32\times 128$, and the number of split blocks is 4. Besides, data augmentation methods such as random cropping, flipping, and transformation are used.
During training, the input image size is randomly adjusted between $320\times320$ and $608\times608$, while 
For testing, the image size is fixed at $544\times544$.  In the all-in-one setting, the model is trained on VM-T for 40 epochs. In the one-by-one setting, we train separate models on VF-HT and VD-HT for 240 epochs. The batch size is set to 16, the optimizer uses SGD\cite{sgd}, and the initial learning rate and the weight decay are set to 0.001 and 0.0005, respectively.
\section{Results and Discussion}
\label{app_3}
\begin{table}[t]
\centering
\caption{The performance of CPA-Enhancer under different number of tasks. In the first column, "Tasks" refers to the number of tasks (degradations).}
\begin{tabular}{@{}c|ccccc|cccccc@{}}
\toprule
\multirow{2}{*}{\textbf{\begin{tabular}[c]{@{}c@{}}Tasks\end{tabular}}} & \multicolumn{5}{c|}{\textbf{Training Datasets}} & \multicolumn{6}{c}{\textbf{Testing Datasets}}     \\
& \textbf{VnA-T} & \textbf{VF-T} & \textbf{VD-T} & \textbf{VS-T} & \textbf{VR-T} & \textbf{VF} & \textbf{VD} & \textbf{VS} & \textbf{VR} & \textbf{RTTS} & \textbf{ExDarkA} \\ \midrule
\multirow{6}{*}{2}                                                                   
& \ding{51}  & \ding{51} & \ding{51}  & \ding{55}  &\ding{55} & 84.12       & 82.04       & -           & -           & 61.63         & 56.83            \\
& \ding{51}&\ding{51}& \ding{55} &\ding{51}&\ding{55}         & 83.94       & -           & 83.11       & -           & 61.53         & -                \\
& \ding{51}& \ding{51} &\ding{55}  & \ding{55} & \ding{51}    & 83.95       & -           & -           & 82.97       & 61.55         & -                \\
& \ding{51}& \ding{55} & \ding{51} & \ding{51} &\ding{55}     & -           & 81.79       & 83.06       & -           & -             & 56.74            \\
& \ding{51} & \ding{55}& \ding{51}&\ding{55}& \ding{51}       & -           & 81.81       & -           & 83.00       & -             & 56.77            \\
& \ding{51}& \ding{55} & \ding{55} &\ding{51} &\ding{51}      & -           & -           & 83.25       & 83.06       & -             & -                \\ \midrule
\multirow{4}{*}{3}  
& \ding{51}&\ding{55} & \ding{51}& \ding{51} & \ding{51}      & -           & 81.47       & 83.01       & 82.77       & -             & 56.64            \\
& \ding{51}& \ding{51} &\ding{55}  &\ding{51} & \ding{51}     & 83.79       & -           & 82.98       & 82.76       & 61.37         & -                \\
& \ding{51}    & \ding{51}& \ding{51}& \ding{55} &\ding{51}   & 83.83       & 81.59       & -           & 82.69       & 61.48         & 56.69            \\
& \ding{51} &\ding{51}& \ding{51}& \ding{51} & \ding{55}     & 83.81       & 81.56       & 82.95       & -           & 61.44         & 56.68            \\ \midrule
\rowcolor{gray!20} \multirow{1}{*}{4}        
& \ding{51}  & \ding{51} & \ding{51}& \ding{51} & \ding{51}   & 83.32       & 80.72       & 82.35       & 82.30       & 61.08         & 56.35            \\ \bottomrule
\end{tabular}
\label{tab:combination}
\end{table}
\begin{table}[t]
\centering

\caption{Impact of the initial prompt size}
\begin{tabular}{@{}c>{\centering\arraybackslash}m{1cm}>{\centering\arraybackslash}m{1cm}>{\centering\arraybackslash}m{1cm}>{\centering\arraybackslash}m{1cm}>{\centering\arraybackslash}m{1cm}>{\centering\arraybackslash}m{1.2cm}>{\centering\arraybackslash}m{1.5cm}@{}}
\toprule
\textbf{Size} & \textbf{VnA}& \textbf{VF}& \textbf{VD}& \textbf{VS}& \textbf{VR}   & \textbf{RTTS}  & \textbf{ExDarkA} \\ \midrule
$64 \times 16\times 16$ &83.25 &82.61&79.04 &81.97  & 81.58 & 55.83 & 59.78           \\
\rowcolor{gray!20} $128 \times 32\times 32$   &84.40&83.32&80.72&82.35        & 82.30 & 56.35 & 61.08  \\
$256 \times 64 \times 64 $   &84.32  &83.34 &81.09& 82.41  & 82.37           & 56.88 & 61.36  \\ \bottomrule
\end{tabular}
\label{tab:prompt_size}
\end{table}
\begin{figure}[h]
  \centering
  \begin{subfigure}[b]{0.45\textwidth}
    \includegraphics[width=\textwidth]{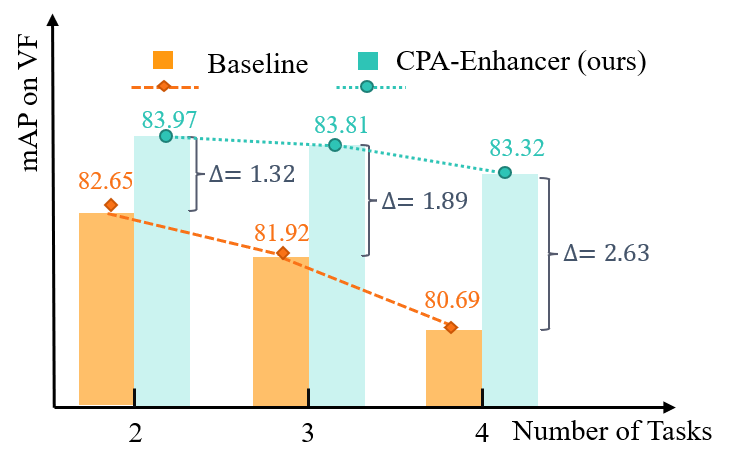}
    \caption{Average preformance on VF.}
  \end{subfigure}
  \begin{subfigure}[b]{0.45\textwidth}
    \includegraphics[width=\textwidth]{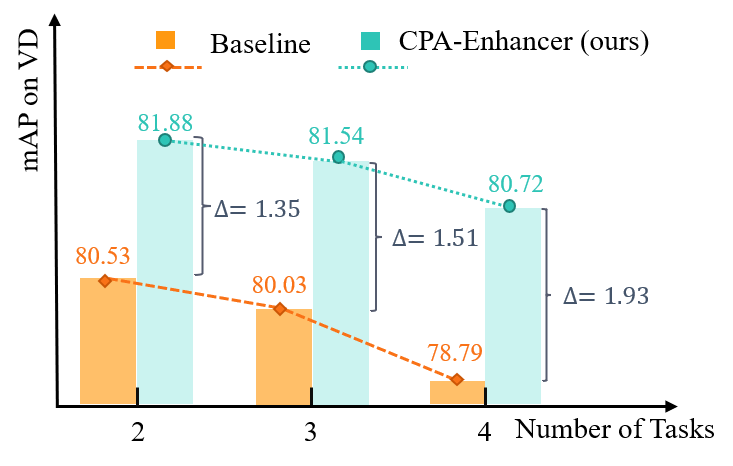}
    \caption{Average preformance on VD.}
  \end{subfigure}
  \begin{subfigure}[b]{0.45\textwidth}
    \includegraphics[width=\textwidth]{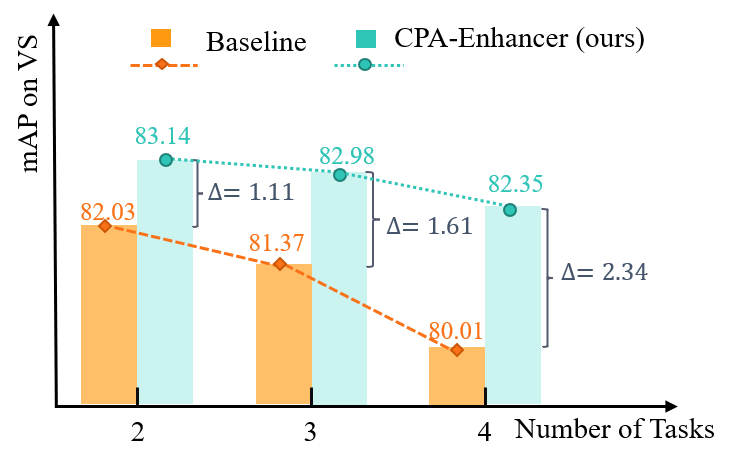}
    \caption{Average preformance on VS.}
  \end{subfigure}
  \begin{subfigure}[b]{0.45\textwidth}
    \includegraphics[width=\textwidth]{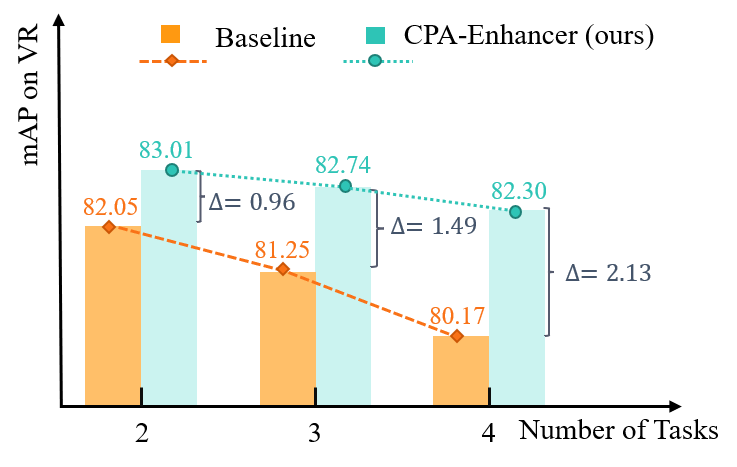}
    \caption{Average preformance on VR.}
  \end{subfigure}
  \caption{Impact of the number of tasks.}
  \label{fig:app_tasknum}
\end{figure}
\begin{table}[h]
\centering
\small
\begin{minipage}[t]{0.48\textwidth}
\centering
\small
\caption{Quantitative comparisons on the ExDark datasets.}
\label{tab:fea_1}
\begin{tabular}{@{}cc@{}}
\toprule
\textbf{Methods}    & \textbf{ExDark} \\ \midrule
YOLOv3\cite{yolov3}             & 43.02            \\
MS-DAYOLO\cite{ms-dayolo}           & 44.25            \\
DAYOLO\cite{dayolo}              & 44.62            \\
DSNet\cite{multitask-dsnet}              & 45.31            \\
MAET\cite{multitask-aet}                & 47.10            \\
IA-YOLO\cite{ia-yolo}             & 49.53            \\
DE-YOLO\cite{denet}             & 51.51            \\
FeatEnHancer\cite{featenhancer}        & \textcolor{blue}{53.70}            \\
\rowcolor{gray!20} CPA-Enhancer (ours) & \textcolor{red}{57.32}            \\ \bottomrule
\end{tabular}
\end{minipage}
\begin{minipage}[t]{0.48\textwidth}
\centering
\small
\caption{Quantitative comparisons on the DARK FACE dataset.}
\label{tab:fea_2}
\begin{tabular}{@{}cc@{}}
\toprule
\textbf{Method}       & \textbf{DARK FACE} \\ \midrule
RetinaNet\cite{retina}     & 47.3      \\
RAUS\cite{raus}         & 42.1      \\
KinD\cite{kind}         & 47.2      \\
EnlightenGAN\cite{enlightengan}        & 45.1      \\
MBLLEN\cite{mbllen}       & 47.1      \\
Zero-DCE\cite{zero-dce}     & \textcolor{blue}{47.4}      \\
MAET\cite{multitask-aet}         & 44.3      \\
FeatEnhancer\cite{featenhancer} & 47.2      \\
\rowcolor{gray!20} CPA-Enhancer (ours)         & \textcolor{red}{47.8}      \\ \bottomrule
\end{tabular}
\end{minipage}
\end{table}
\subsection{Impact of the Numbers of Tasks} 
\label{app_31}
We train CPA-Enhancer and the baseline YOLOv3\cite{yolov3} under the all-in-one setting on different combinations of training datasets. As reported in \cref{tab:combination}, the increasing number of tasks precipitates a concomitant decline in the performance of CPA-Enhancer. \cref{fig:app_tasknum} shows the average performance of CPA-Enhancer and YOLOv3 under different number of tasks. It can be observed that the performance degradation of YOLOv3 on four datasets deteriorates more significantly with the increasing number of tasks. Thanks to the introduction of CoT prompts, CPA-Enhancer can dynamically adjust its enhancement strategies based on the degradation type. This adaptability allows CPA-Enhancer to better maintain its performance across different degradation types, resulting in a smaller performance decrease compared to YOLOv3 as the number of tasks (degradation types) increases.

\subsection{Impact of the Initial Prompt Size}
\label{app_32}
\cref{tab:prompt_size} lists the performance of CPA-Enhancer with different sizes of the initial prompt. As the initial prompt size increases, the detection performance becomes better. However, when the size is $256\times 64\times 64$, 2M more parameters are introduced, which only brings a limited improvement. To trade off performance and efficiency, we choose a size of $128\times 32\times 32$ in our default setting.
\subsection{Illustration of SPB}
\label{app_spb}
\begin{figure*}[t]
\centering
	\includegraphics[width=1\textwidth]{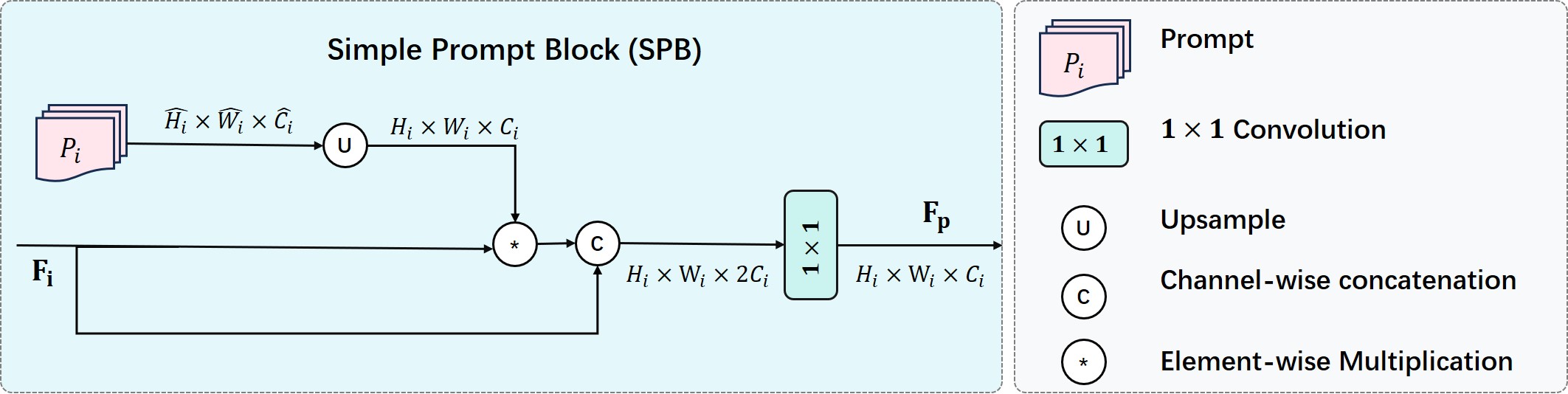}
	\caption{Illustration of SPB.}
        \label{fig:spb}
\end{figure*}

To evaluate the effectiveness of our proposed content-driven prompt block(CPB), we design a simple prompt block (SPB) for comparison. Here, we illustrates SPB in \cref{fig:spb}. 
\subsection{Comparison with FeatEnHancer in the Low-light Condition}
\label{app_33}
FeatEnHancer\cite{featenhancer} is a general-purpose plug-and-play module designed for the low-light condition, which achieves state-of-the-art results in several downstream vision tasks. To further demonstrate the effectiveness of our method under a single degradation condition, we compare CPA-Enhancer with FeatEnHancer by conducting the following experiments, including dark object detection and face detection.

\begin{table}[t]
\centering
\caption{Quantitative comparisons on the ACDC test set}
\resizebox{1\textwidth}{!}{
\begin{tabular}{@{}ccccccccccccccccccccc@{}}
\toprule
\textbf{Method}                                                          
& \rotatebox{90}{road} & \rotatebox{90}{sidew.} & \rotatebox{90}{build.}& \rotatebox{90}{wall.} & \rotatebox{90}{fence.} & \rotatebox{90}{pole.} & \rotatebox{90}{light.} & \rotatebox{90}{sign.} & \rotatebox{90}{veget.} & \rotatebox{90}{terrain.} & \rotatebox{90}{sky.} & \rotatebox{90}{person.} & \rotatebox{90}{rider.} & \rotatebox{90}{car.} & \rotatebox{90}{truck.} & \rotatebox{90}{bus.} & \rotatebox{90}{train.} & \rotatebox{90}{motorc.} & \rotatebox{90}{bicycle.} & \rotatebox{90}{\textbf{mean}} \\ \midrule
\textbf{MGCDA}\cite{mgcda}
&73.4&28.7&69.9&19.3&26.3&36.8&53.0&53.3&75.4&32.0&84.6&51.0&26.1&77.6&43.2&45.9&53.9&32.7&41.5 &48.7\\
\textbf{DANNet}\cite{dannet} &84.3&54.2&77.6&38.0&30.0&18.9&41.6&35.2&71.3&39.4&86.6&48.7&29.2&76.2&41.6&43.0&58.6&32.6&43.9&50.0 \\
\textbf{DANIA}\cite{dania} &88.4&60.6&81.1&37.1&32.8&28.4&43.2&42.6&77.7&50.5&90.5&51.5&31.1&76.0&37.4&44.9&64.0&31.8& 46.3&53.5 \\
\textbf{DAFormer}\cite{seg-daformer} 
& 58.4&51.3&84.0&42.7&35.1&50.7&30.0&57.0&74.8&52.8&51.3&58.3&32.6&82.7&58.3&54.9&82.4&44.1&50.7&55.4\\
\textbf{Refign}\cite{refign} 
&\textcolor{blue}{89.5}&\textcolor{blue}{63.4}&\textcolor{blue}{87.3}&\textcolor{blue}{43.6}&\textcolor{blue}{34.3}&\textcolor{blue}{52.3}&\textcolor{blue}{63.2}&\textcolor{blue}{61.4}&\textcolor{blue}{86.9}&\textcolor{blue}{58.5} &\textcolor{blue}{95.7}&\textcolor{blue}{62.1}&\textcolor{blue}{39.3}&\textcolor{blue}{84.1}&\textcolor{blue}{65.7}&\textcolor{red}{71.3} &\textcolor{blue}{85.4}&\textcolor{blue}{47.9}&\textcolor{blue}{52.8}&\textcolor{blue}{65.5} \\
\rowcolor{gray!20} \textbf{Ours (S)}
 &\textcolor{red}{95.2}&\textcolor{red}{80.4}&\textcolor{red}{91.2}&\textcolor{red}{62.3}&\textcolor{red}{53.8}&\textcolor{red}{61.8}&\textcolor{red}{74.8}&\textcolor{red}{72.3}&\textcolor{red}{89.7}&\textcolor{red}{68.0}&\textcolor{red}{96.7}&\textcolor{red}{69.2}&\textcolor{red}{41.0} &\textcolor{red}{88.6}&\textcolor{red}{70.3}&\textcolor{blue}{59.2}&\textcolor{red}{89.6}&\textcolor{red}{49.0}&\textcolor{red}{58.5}&\textcolor{red}{72.2}   \\ \bottomrule
\end{tabular}
}
\label{tab:seg}
\end{table}
\subsubsection{Dark Object Detection on ExDark.}
Following FeatEnHancer, we train our CPA-Enhancer on VD-HT for 40 epochs and test its performance on the ExDark datasets\cite{exdark}. The other settings and hyperparameters are identical to the experiments explained in \cref{app_2}. As shown in \cref{tab:fea_1}, CPA-Enhancer significantly surpasses the previous state-of-the-art method (FeatEnHancer) by a margin of $3.62\%$ mAP$_{50}$. 
\subsubsection{Face Detection on DARK FACE.}
\begin{figure*}[t]
\centering
	\includegraphics[width=1\textwidth]{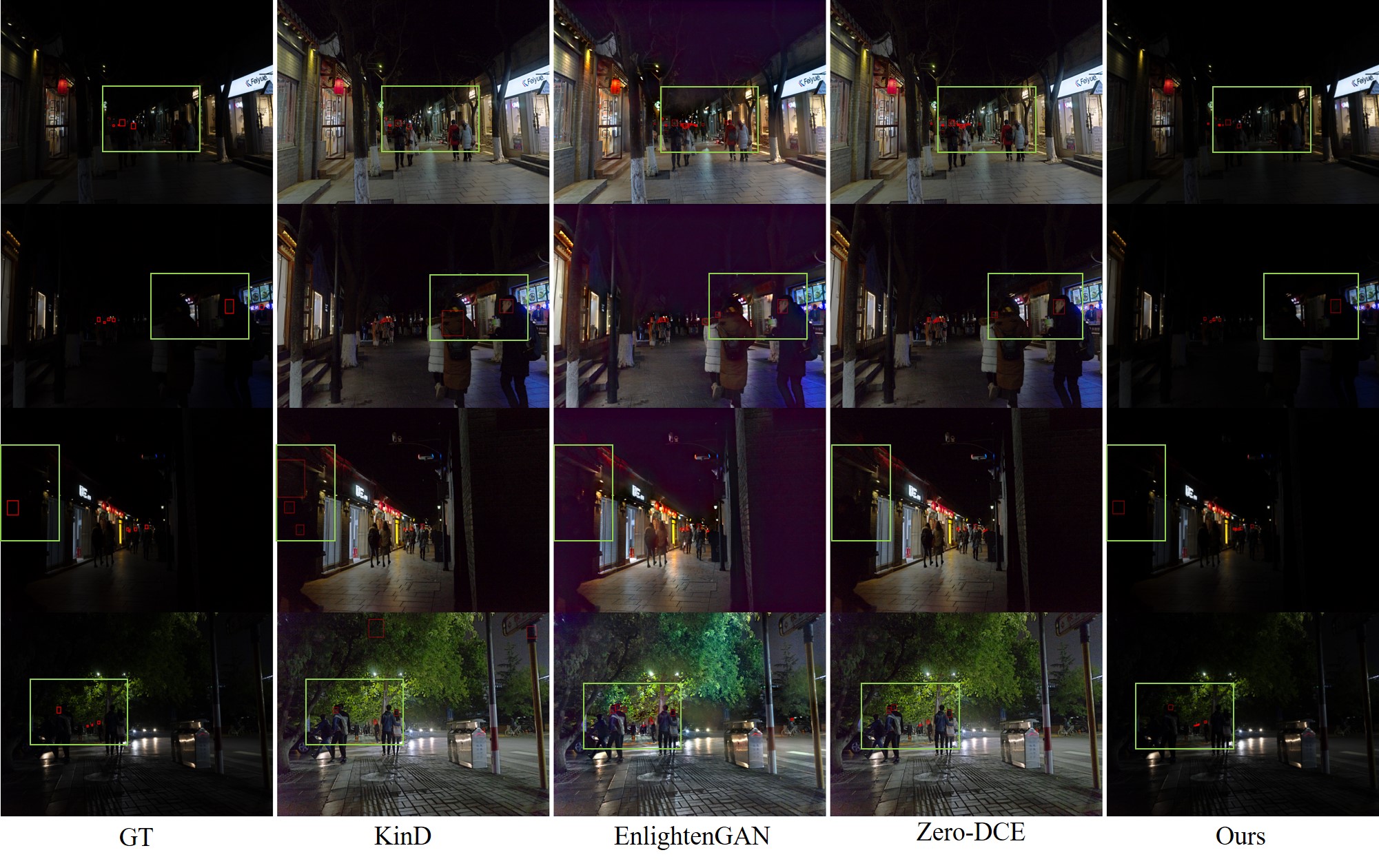}
	\caption{Qualitative comparisons on the DARK FACE dataset. Zoom in on the green annotation boxes to better observe the differences.}
        \label{fig:dark_face}
\end{figure*}

The DARK FACE dataset\cite{darkface} provides 6,000 real-world low-light images,  where human faces are labeled with annotated boxes. 
We follow the same experimental setup as FeatEnhancer for fair comparisons. Specifically, RetinaNet\cite{retina} is adopted as our baseline detector, which works in an end-to-end learning fashion with our proposed CPA-Enhancer. All the images are resized to $1500\times 1000$, SGD optimizer\cite{sgd} with an initial learning rate of 0.001 is utilized. The batch size is set to 8 during the training. 
The DARK FACE dataset is split into training (5400 images) and validation (600 images) sets and we apply mAP$_{50}$ as the evaluation metrics. 
We compare our proposed CPA-Enhancer with several low-light enhancement methods (KIND\cite{kind}, RAUS\cite{raus}, EnlightenGAN\cite{enlightengan}, MBLLEN\cite{mbllen}, Zero-DCE\cite{zero-dce}), and dark object detection methods MAET\cite{multitask-aet}, and FeatEnHancer\cite{featenhancer}. Displayed in \cref{tab:fea_2}, our proposed CPA-Enhancer achieves a mAP$_{50}$ of 47.8\% on the DARK FACE dataset, setting the new state-of-the-art.
The visual comparison in \cref{fig:dark_face} illustrates that while the low-light enhancement methods can enhance the image exposure to some extent, they unfortunately generate more false negatives (FN) and false positives (FP). On the contrary, our proposed CPA-Enhancer successfully produces more accurate results in face detection, demonstrating the strong adaptability and robustness of our method in the low-light condition.
\begin{figure*}[t]
\centering
	\includegraphics[width=1\textwidth]{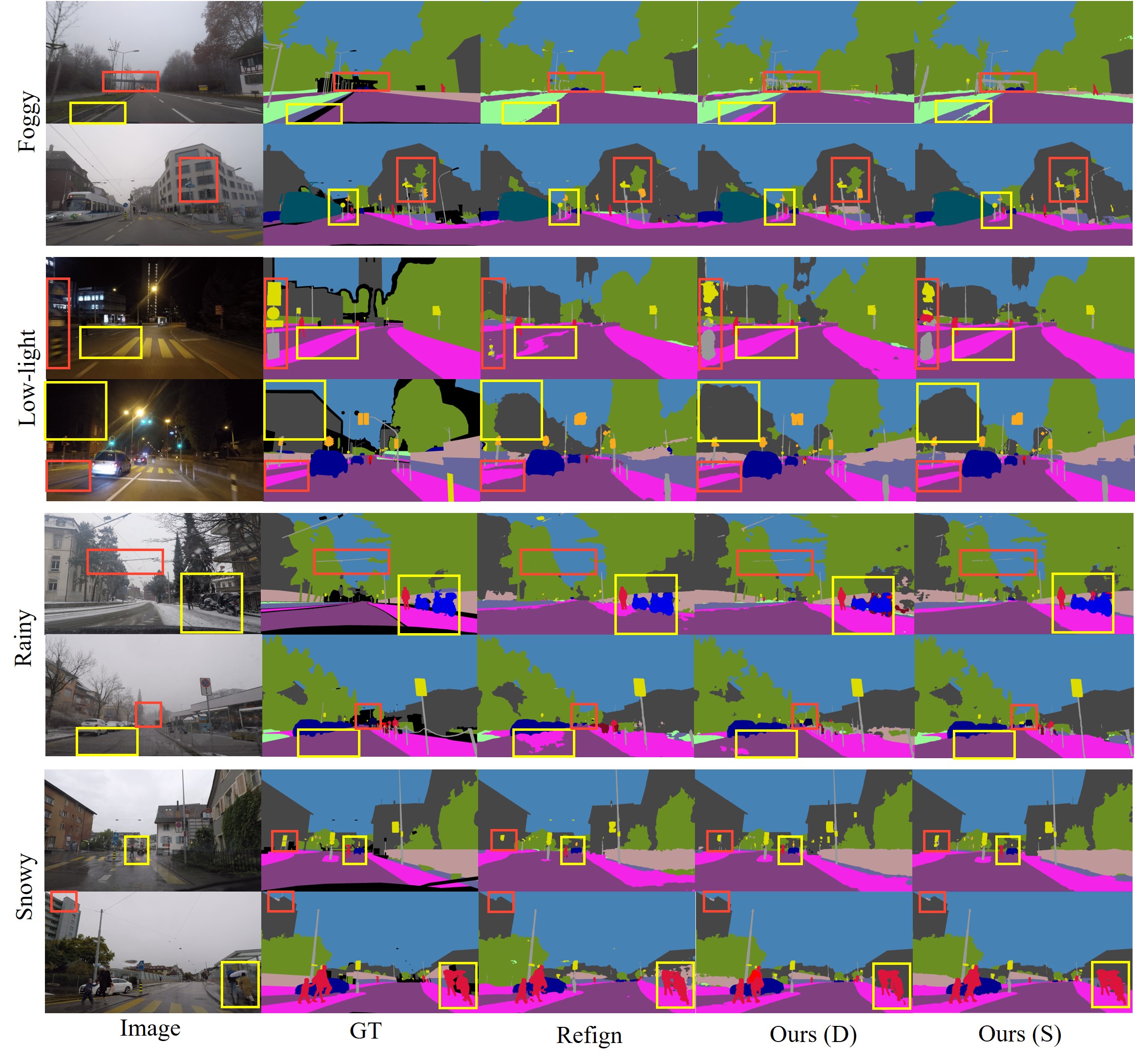}
	\caption{Qualitative comparisons of semantic segmentation on the ACDC validation set. Zoom in on the colored annotation boxes to better observe the differences.}
        \label{fig:app_seg}
\end{figure*}
\subsection{Semantic Segmentation on ACDC}
\label{app_34}
The ACDC dataset\cite{acdc} comprises 1600 training images, 406 validation images, and 2000 test images, all evenly distributed across four weather conditions: fog, night, rain, and snow. 
We cascade our proposed CPA-Enhancer with two basic segmentation models, DeepLabv3+\cite{seg-deeplabv3} and Segformer\cite{segformer}, and labeled them as Ours(D) and Ours(S) respectively. During the training process, AdamW optimizer\cite{adamw} is used with a learning rate of 0.00006. The weight decay is set to 0.01. The images are resized to $2048\times 1024$ for the training with a batch size of 4. The Mean Intersection over Union (mIoU) metric is employed for presenting the segmentation results.

We compare our method with several unsupervised domain adaptation-based methods, incuding MCCDA\cite{mgcda}, DANNet\cite{dannet}, DANIA\cite{dania}, DAFormer\cite{seg-daformer} (using SegFormer\cite{segformer}), and Refign (based on DAFormer\cite{seg-daformer}).  
We present quantitative comparisons on the ACDC test set, and corresponding results are provided in \cref{tab:seg}. As expected, our model continues to achieve the best result with a mIoU of 72.2\%, substantially outperforming the previous best result (Refign\cite{refign}) by 6.7 points. For convincing evidence, we further show the visual quality comparisons in \cref{fig:app_seg}. Our CPA-Enhancer consistently outperforms Refine in terms of generating more precise segmentation maps under different degradation types, indicating the versatility and robustness of our approach in handling visual tasks under multiple unknown degradations. 

\begin{figure*}[t]
\centering
	\includegraphics[width=1\textwidth]{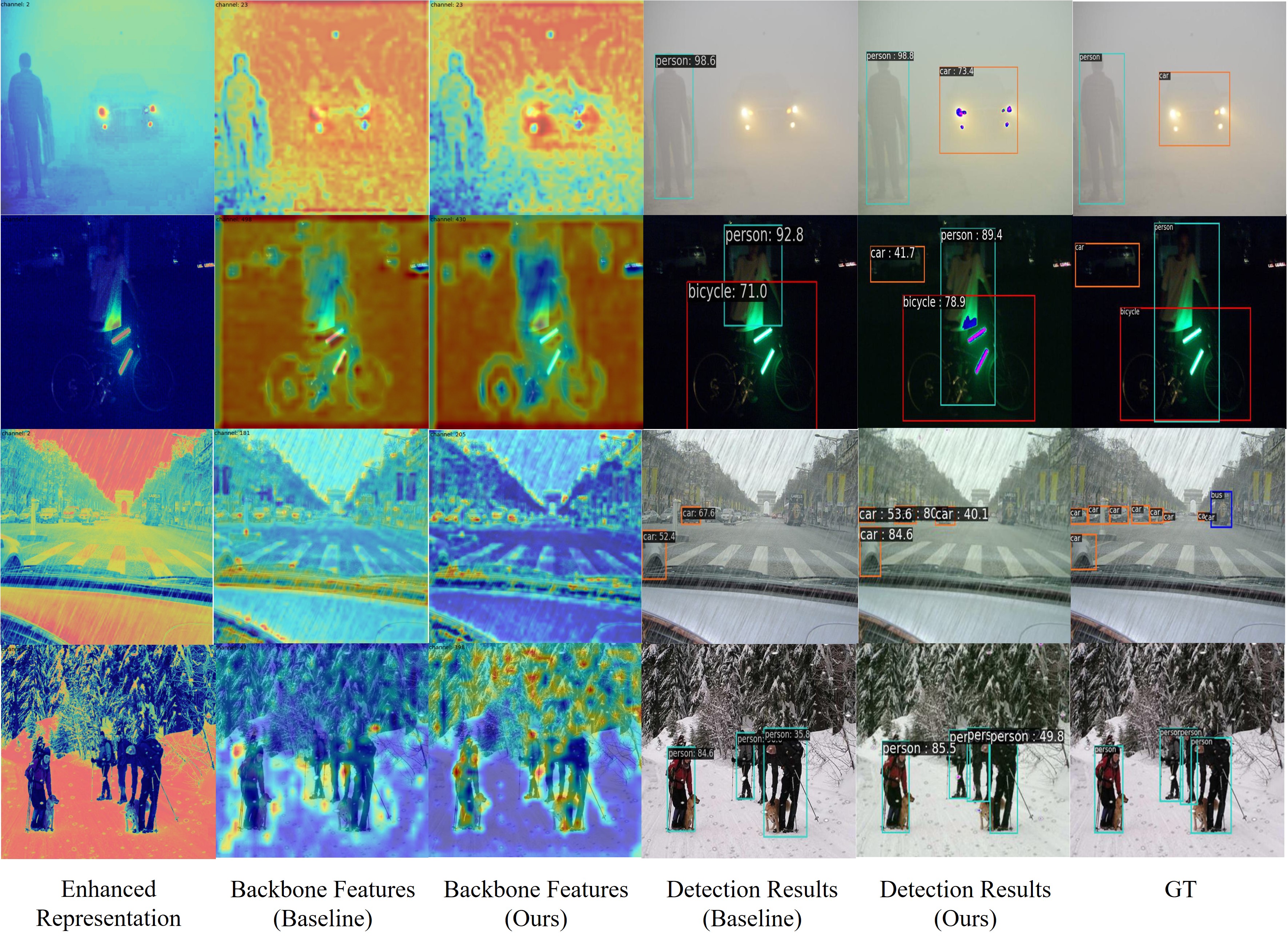}
	\caption{Visual enhanced representations and backbone features.}
        \label{fig:heatmap}
\end{figure*}
\subsection{Additional Qualitative Results for the All-in-one Setting}
\label{app_35}
\begin{figure*}[p]
\centering
	\includegraphics[width=1\textwidth]{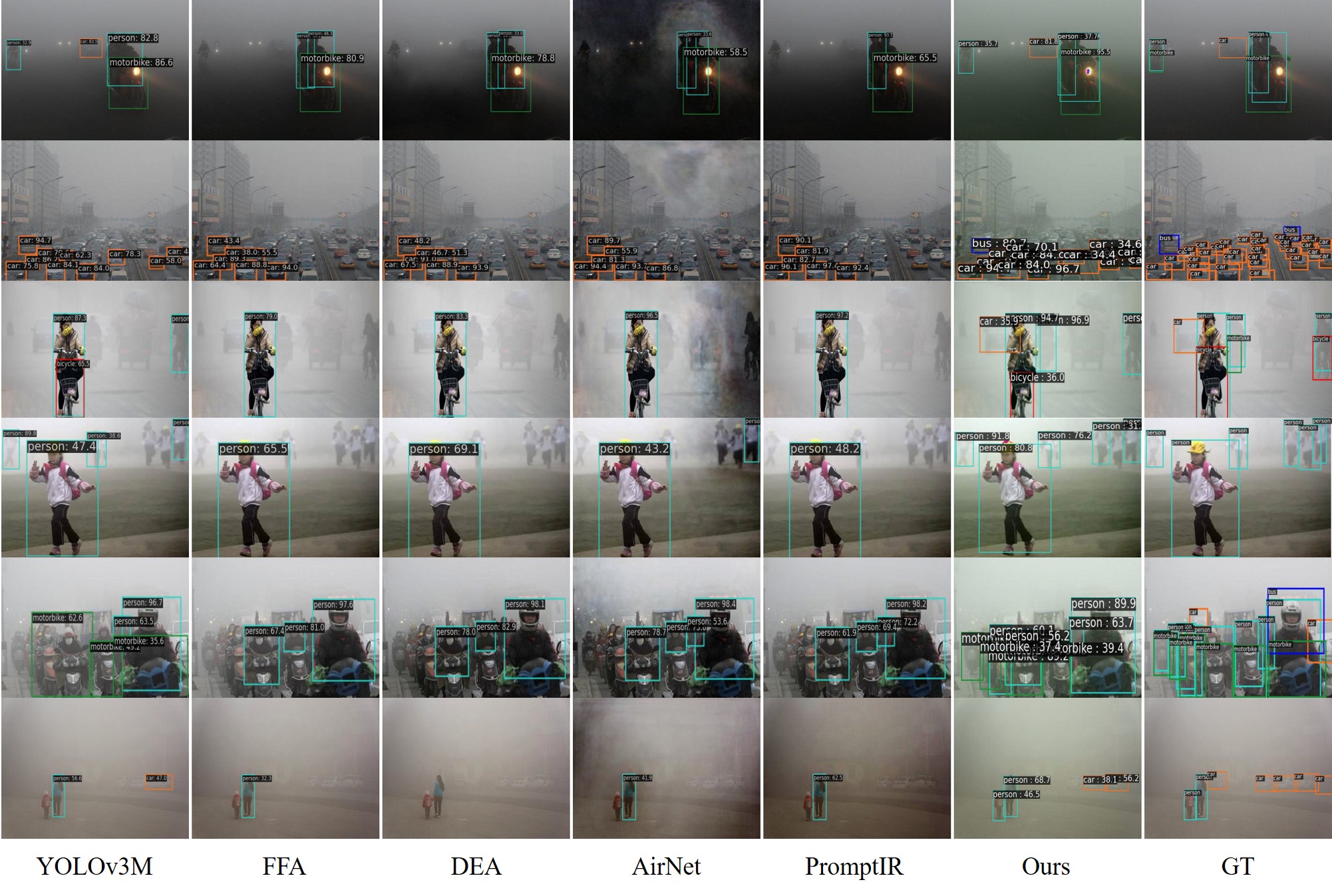}
	\caption{Qualitative comparisons of object detection on RTTS in the all-in-one setting.}
        \label{fig:app_allinone_fog}
\end{figure*}

\begin{figure*}[p]
\centering
	\includegraphics[width=1\textwidth]{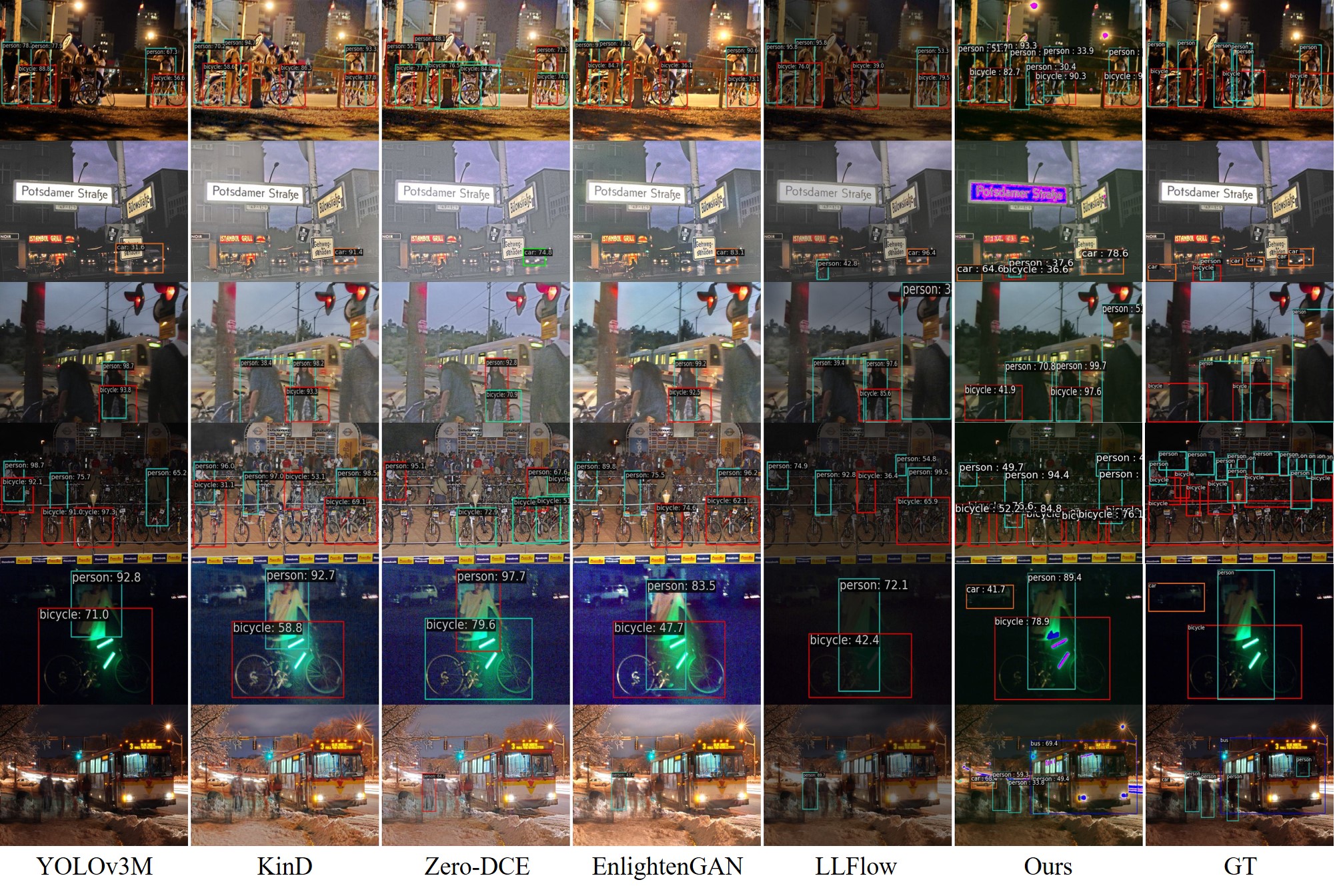}
	\caption{Qualitative comparisons of object detection on ExDark in the all-in-one setting.}
        \label{fig:app_allinone_night}
\end{figure*}

\begin{figure*}[p]
\centering
	\includegraphics[width=1\textwidth]{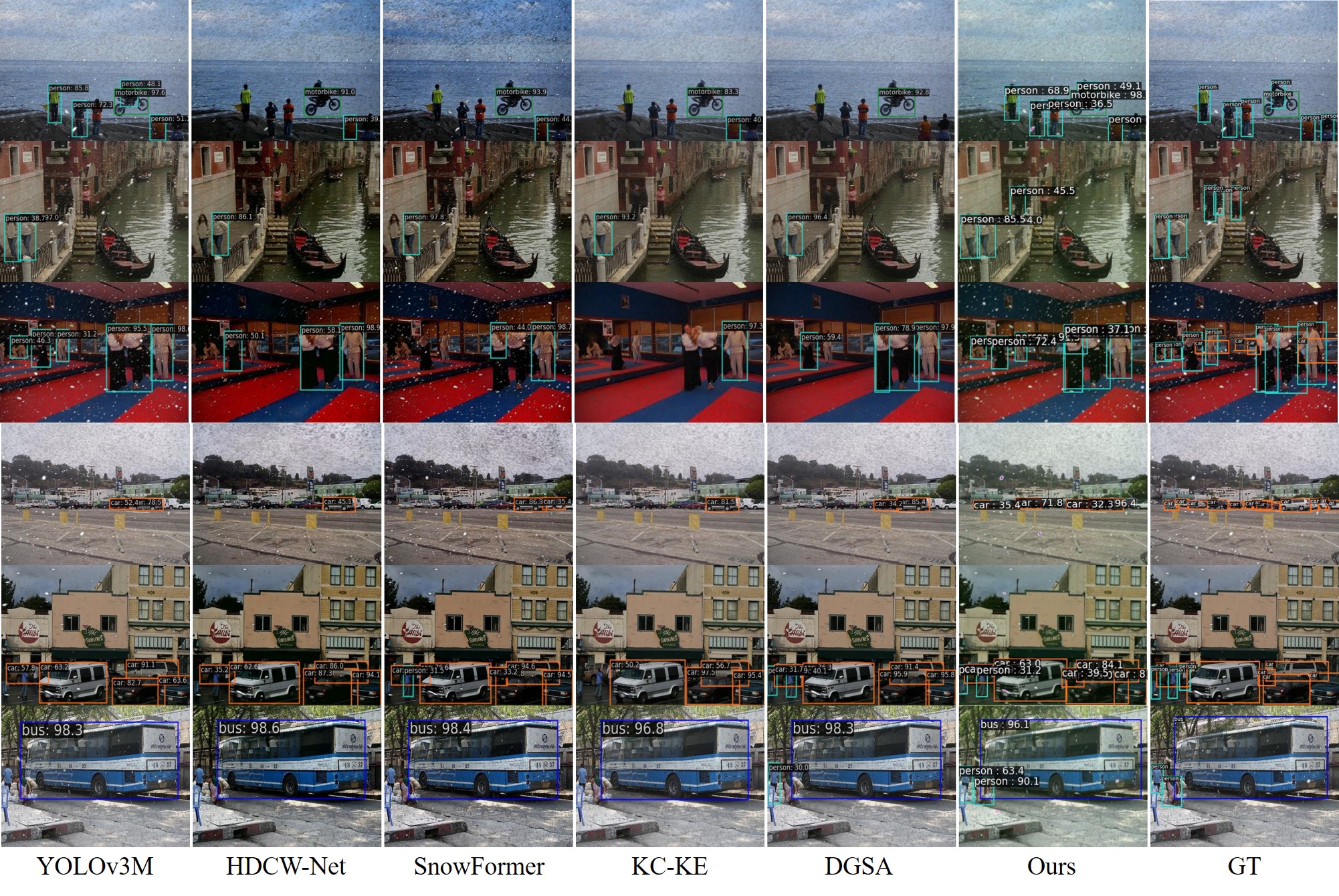}
	\caption{Qualitative comparisons of object detection on VS in the all-in-one setting.}
        \label{fig:app_allinone_snow}
\end{figure*}

\begin{figure*}[p]
\centering
	\includegraphics[width=1\textwidth]{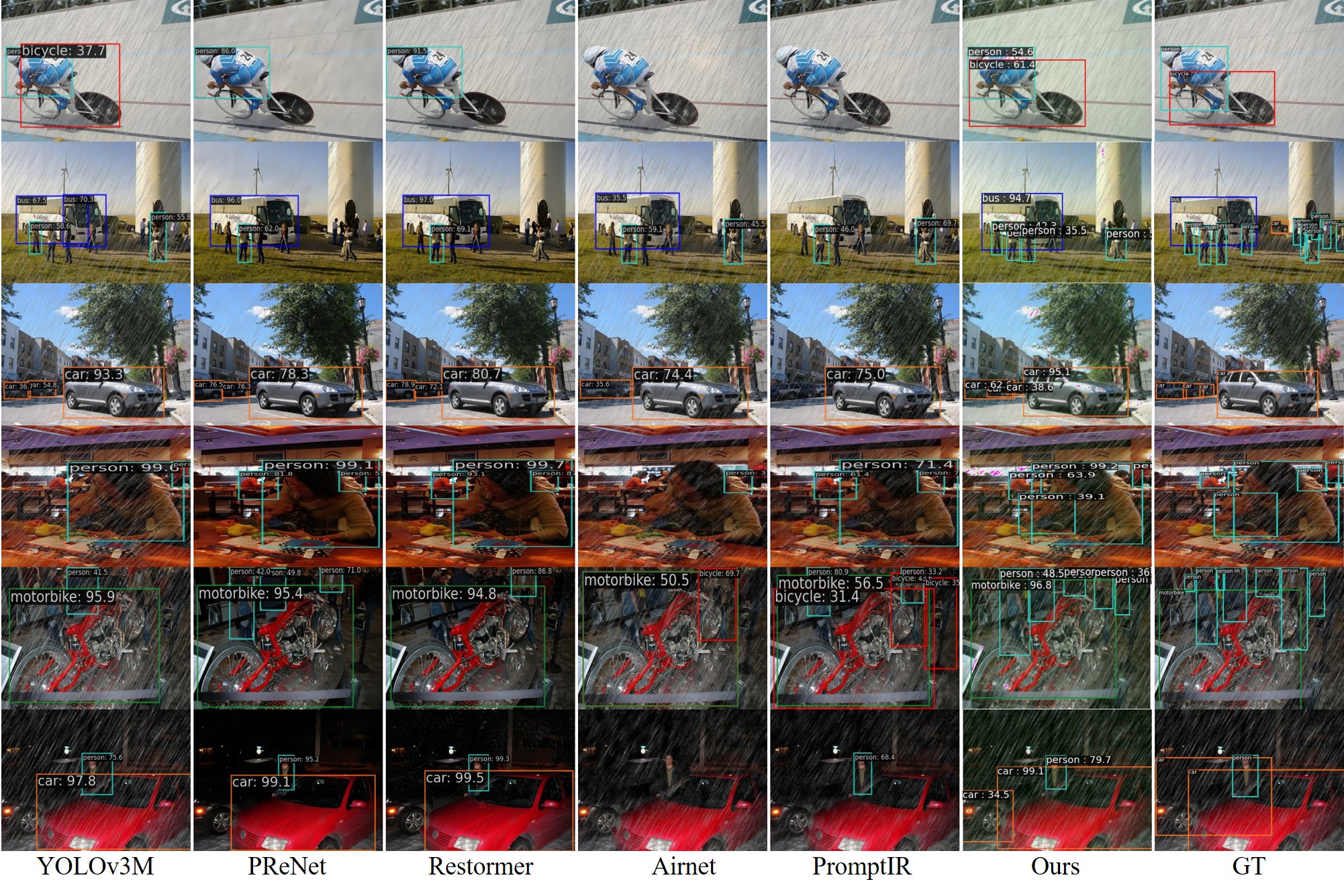}
	\caption{Qualitative comparisons of object detection on VR in the all-in-one setting.}
        \label{fig:app_allinone_rain}
\end{figure*}

To further elucidate the effectiveness of CPA-Enhancer under different types of degradations, we provide enhanced representations and backbone features of our CPA-Enhancer and the baseline YOLOv3\cite{yolov3}. As shown in \cref{fig:heatmap}, the well-enhanced representations of our method lead to better feature extraction and representation, thereby improving the detection performance.
Additionally, we also present more qualitative results for the all-in-one setting. 
As shown in \cref{fig:app_allinone_fog}, we compare CPA-Enhancer with several defog methods (FFA\cite{ffa}, DEA\cite{dea}, AirNet\cite{airnet}, and PromptIR\cite{prompt-ir}) on the RTTS dataset\cite{rtts}. 
In \cref{fig:app_allinone_night}, we juxtapose CPA-Enhancer with several defog methods (KinD\cite{kind}, Zero-DCE\cite{zero-dce}, EnlightenGAN\cite{enlightengan}, and LLFlow\cite{llflow}) on the ExDark dataset\cite{exdark}.
As depicted in \cref{fig:app_allinone_snow}, we also evaluate the performance of CPA-Enhancer alongside several existing desnow methods (HDCW-Net\cite{snow-hdcw}, SnowFormer\cite{snowformer}, KC-KE\cite{snow-twostage}, and DGSA\cite{snow-dual}) on the VS dataset. In \cref{fig:app_allinone_rain}, we further contrast CPA-Enhancer against several derain methods (PReNet\cite{prenet}, Restoremr\cite{restormer}, AirNet\cite{airnet}, and PromptIR\cite{prompt-ir}). 
Attributed to the design of the CoT prompts, our CPA-Enhancer performs exceptionally well in various unknown degradations, demonstrating stronger adaptability and detection accuracy compared to other methods. 

\section{Detailed Efficiency Analysis}
\label{app_4}
Furthermore, We compare our method with the previous state-of-the-art dehazing\cite{ffa,dea,airnet,promptir} (see \cref{tab:eff_1}), low-light enhancement\cite{kind, zero-dce, enlightengan, llflow} (see \cref{tab:eff_2}), desnowing\cite{snow-hdcw,snowformer,snow-twostage,snow-dual} (see \cref{tab:eff_3}), and deraining\cite{prenet,restormer,airnet,prompt-ir} (see \cref{tab:eff_4}) methods in terms of model complexity and parameters. The GFLOPs are calculated in
$256\times 256$ resolution. 
Our method achieves high performance on all the testing datasets with relatively low model complexity, utilizing 3.43 million parameters and 12.93 GFLOPs. 
However, it is noteworthy that the majority of parameters originate from the transformer block in our proposed CPB. We anticipate that replacing the transposed attention with a more optimized attentional scheme\cite{flatten} or applying retrieval augmented method\cite{selfrag, retrieval, raisf} to improve CPA-Enhancer performance, could lead to further parameter reduction in our CPA-Enhancer. This avenue remains open for exploration in future research. 
\begin{table}[t]
\centering
\small
\begin{minipage}[t]{0.51\textwidth}
\centering
\small
\caption{Efficiency comparisons with defog methods.}
\label{tab:eff_1}
\resizebox{1\textwidth}{!}{
\begin{tabular}{@{}ccccc@{}}
\toprule
\textbf{Method} & \textbf{\#Param} & \textbf{\#GFLOPs} & \textbf{VF} & \textbf{RTTS} \\ \midrule
FFA             & 4.46M            & 287.53G           & 75.38       & 49.46         \\
DEA             & 3.56M            & 34.04G            & 81.02       & 50.10         \\
AirNet          & 44M              & 12.16G            & 76.2        & 48.32         \\
PromptIR        & 32.11M           & 158.15G           & 75.98       & 50.64         \\
\rowcolor{gray!20} Ours            & 3.43M            & 12.93G            & 83.32       & 56.35         \\ \bottomrule
\end{tabular}
}
\end{minipage}
\begin{minipage}[t]{0.45\textwidth}
\centering
\small
\caption{Efficiency comparisons with desnow methods.}
\label{tab:eff_3}
\resizebox{1\textwidth}{!}{
\begin{tabular}{@{}cccc@{}}
\toprule
\textbf{Method}     & \textbf{\#Param} & \textbf{\#GFLOPs} & \textbf{VS}    \\ \midrule
HDCW-Net   & 6.99M   & 9.78G    & 77.82 \\
SnowFormer & 8.38M   & 19.44G   & 79.46 \\
KC-KE      & 31.35M  & 41.58G   & 79.63 \\
DGSA       & 2.07M   & 133.69G  & 78.19 \\
\rowcolor{gray!20} Ours       & 3.43M   & 12.93G   & 82.35 \\ \bottomrule
\end{tabular}}
\end{minipage}

\end{table}
\begin{table}[t]
\centering
\small
\begin{minipage}[t]{0.55\textwidth}
\centering
\small
\caption{Efficiency comparisons with low-light enhancement methods.}
\label{tab:eff_2}
\resizebox{1\textwidth}{!}{
\begin{tabular}{@{}ccccc@{}}
\toprule
\textbf{Method} & \textbf{\#Param} & \textbf{\#GFLOPs} & \textbf{VD} & \textbf{ExDarkA} \\ \midrule
KinD            & 7.36M            & 0.24G             & 74.28       & 52.78            \\
Zero-DCE        & 79K              & 5.20G             & 76.64       & 53.52            \\
EnlightenGan    & 8.64M            & 1150.87G          & 76.42       & 53.16            \\
LLFlow          & 1.85M            & 103.50G           & 77.08       & 53.90            \\
\rowcolor{gray!20}  Ours            & 3.43M            & 12.93G            & 80.72       & 61.08            \\ \bottomrule
\end{tabular}
}
\end{minipage}
\begin{minipage}[t]{0.40\textwidth}
\centering
\small
\caption{Efficiency comparisons with derain methods.}
\label{tab:eff_4}
\resizebox{1\textwidth}{!}{
\begin{tabular}{@{}cccc@{}}
\toprule
\textbf{Method} & \textbf{\#Param} & \textbf{\#GFLOPs} & \textbf{VR} \\ \midrule
PReNet          & 168K             & 66.24G            & 81.32       \\
Restormer       & 26.10M           & 140.99G           & 81.38       \\
AirNet          & 44M              & 12.16G            & 72.14       \\
PromptIR        & 32.11M           & 158.15G           & 73.78       \\
\rowcolor{gray!20} Ours            & 3.43M            & 12.93G            & 82.30       \\ \bottomrule
\end{tabular}
}
\end{minipage}
\end{table}
\end{document}